\documentclass[authorversion, screen=true]{acmart}
\AtBeginDocument{%
  \providecommand\BibTeX{{%
    \normalfont B\kern-0.5em{\scshape i\kern-0.25em b}\kern-0.8em\TeX}}}

\setcopyright{none}

\usepackage{soul}
\usepackage{url}
\usepackage{cleveref}
\usepackage[utf8]{inputenc}
\usepackage{graphicx}
\usepackage{soul}
\usepackage{array}
\usepackage{booktabs}
\usepackage{multicol}
\usepackage{todonotes} 
\usepackage{subcaption} 
\usepackage{pgf} 
\usepackage{wrapfig} 
\newcolumntype{R}[1]{>{\raggedleft\arraybackslash}b{#1}}
\newcolumntype{L}[1]{>{\raggedright\arraybackslash}b{#1}}

\DeclareMathOperator*{\E}{\mathbb{E}}

\crefformat{section}{\S#2#1#3} 
\crefformat{subsection}{\S#2#1#3}
\crefformat{subsubsection}{\S#2#1#3}

\begin{document}

\title{Fair Decision-Making for Food Inspections}

%
\author{Shubham Singh}
\email{ssing57@uic.edu}
\author{Bhuvni Shah}
\email{bshah46@uic.edu}
\author{Chris Kanich}
\email{ckanich@uic.edu}
\author{Ian A. Kash}
\email{iankash@uic.edu}
\affiliation{%
  \institution{University of Illinois at Chicago}
  \streetaddress{1200 W Harrison St}
  \city{Chicago}
  \state{Illinois}
  \country{USA}
  \postcode{60607}
}

\begin{abstract}
Data and algorithms are essential and complementary parts of a large-scale decision-making process. However, their injudicious use can lead to unforeseen consequences, as has been observed by researchers and activists alike in the recent past. 
In this paper, we revisit the application of predictive models by the Chicago Department of Public Health to schedule restaurant inspections and prioritize the detection of critical food code violations. We perform the first analysis of the model's fairness to the population served by the restaurants in terms of average time to find a critical violation.
We find that the model treats inspections unequally based on the sanitarian who conducted the inspection and that, in turn, there are geographic disparities in the benefits of the model.
We examine four alternate methods of model training and two alternative ways of scheduling using the model and find that the latter generate more desirable results.
The challenges from this application point to important directions for future work around fairness with collective entities rather than individuals, the use of critical violations as a proxy, and the disconnect between fair classification and fairness in the dynamic scheduling system.
\end{abstract}



\maketitle

\section{Introduction}

The Chicago Department of Public Health (CDPH) issues food safety guidelines and conducts inspections of more than 16,000 food establishments. Through these inspections, CDPH sanitarians educate owners and workers about food safety practices, inspect the premises and practices for safe food handling, and promote a healthy environment for food preparation.
The City of Chicago records each of the food inspections on its public data portal.\footnote{Chicago Data Portal: \url{https://data.cityofchicago.org/Health-Human-Services/Food-Inspections/4ijn-s7e5}} 

As there are a limited number of sanitarians, a natural goal is to use data to prioritize performing the inspections that best protect the public health.  Inspections which identify critical violations of the food code allow conditions posing the highest risk of causing a food-borne illness to be addressed.
Thus, data scientists working for the city and their collaborators trained a machine learning (ML) model to predict the likelihood of an inspection resulting in a critical violation~\cite{schenkjr15a}.  The trained ML model, which we refer to as the \citeauthor{schenkjr15a} model, was used to prioritize food inspections in a simulated study. An evaluation of the model showed that using it to schedule inspections achieves a 7-day improvement in the mean time to detect a critical violation compared to the actual inspection schedule followed by sanitarians ~\cite{schenkjr15a}.

In this paper, we first reexamine the model from a fairness perspective and assess how the improvement gained by employing the model is shared by different parts of the city.
A key driver of geographic variation is that not all sanitarians report critical violations at the same rate, with some citing such violations in less than three percent of inspections while others find them in more than forty percent.
We show that the idiosyncratic behavior of sanitarians coupled with each sanitarian working in a limited portion of the city results in a disparity among how soon critical violations are found in different city regions under the \citeauthor{schenkjr15a} model. In particular, the model prioritizes the restaurants inspected by sanitarians who report a high rate of critical violations. 
As a result, residents of regions of the city where sanitarians cite critical violations at a higher-than-average rate tend to see inspections in their region prioritized at the expense of the other regions.

We also explore approaches to using this data to prioritize food inspections in a fairer way.
Our interventions span two broad classes of techniques: (a) those where we train a new model to predict critical violations in a fairer way and (b) \textit{post-processing} approaches where we use the \citeauthor{schenkjr15a} model as-is but modify the way the model is used to achieve a fairer resource allocation despite the unfair predictions.
We examine four different approaches to training fair models, and find that they mitigate but do not eliminate the geographic unfairness that results when the models are used to schedule inspections.  
We consider two post-processing approaches which adjust the way sanitarian identities are used when scheduling inspections: one censors the sanitarian feature when predicting while the other uses the model to reschedule each sanitarian's inspections rather than globally rescheduling all inspections.  Our results show that these post-processing approaches are more effective than retraining in achieving fair outcomes.

After analyzing the fairness properties of the various approaches, we examine the trade off between efficiency (success in prioritizing inspections with critical violations) and fairness each enables. We find that 
the post-processing approaches enable an attractive trade-off between efficiency and fairness while the fair models are essentially Pareto dominated: each has an alternative that provides approximately the same efficiency while being fairer.

We conclude by discussing three key issues our results raise for future work.  The inspection of restaurants is different from much of the fairness literature in that each entity has many stakeholders, so models based on simple binary protected attributes are not a good fit for some protected groups of interest.  The use of critical violations as a proxy for public health raises important questions about what fairness means in this setting, particularly in light of the heterogeneity across sanitarians.  Finally, we discuss the disconnect between training a model for classification and our goal of achieving fairness in a dynamic scheduling system.


\subsection{Related Work}

Food inspections remain an essential food safety practice to prevent food-borne illnesses. 
However, the variation in practices and guidelines across jurisdictions results in a lack of consistency in local and national food safety levels and the link between inspections and food safety is actively studied. 
One of the first studies done in Seattle-King County found that restaurants with lower food inspection scores\footnote{Sanitarians assigned each inspection a score out of 100, and a lower score indicated more critical violations.} were likely to have more outbreaks than restaurants with higher scores~\cite{irwin_results_1989}. 
Another study done in Miami–Dade County found no correlation between restaurant inspection scores and the outbreak of food-borne illnesses~\cite{cruz_assessment_2001}.  \citeauthor{jones_restaurant_2004} found inconsistencies between criteria for high risk establishments and establishments that resulted in outbreakes through a study in the state of Tennessee and called for a deeper examination into the restaurant inspection system ~\cite{jones_restaurant_2004}.
Today, there are still calls for systemic change in local food safety inspection systems across the country due to the poor predictive power of the current inspection framework~\cite{AshwoodZoeC2021ItRo}.
Technologists have also been exploring how to improve the food inspection process. In attempt to find more predictive features, nontraditional data sources have been explored. Google search activity in an ML-based approach to identify higher risk establishments has shown some promise ~\cite{sadilek_machine-learned_2018}, but the addition of Yelp data did not improve the prediction of inspection scores~\cite{altenburger2019yelp}.


More broadly our work sits at the intersection of two trends in the use of artificial intelligence techniques.  One is the use of predictive analytics and other tools to bring algorithmic decision-making to government operations. We witness a rise of machine learning algorithms being used for the delivery of public services, digitization of court records, and management of government programs~\cite{coglianese2020ai,engstrom2020government}. 
The other is the increased openness of government data. A recent study points out the need for more transparency to counter the public distrust of AI and promote its use for the common good~\cite{knowles_sanction_2021}. Such initiatives help authorities improve their operations and also provides transparency to their decision-making process. Tournaments on platforms like Kaggle have been used encourage public involvement in civic model creation ~\cite{10.1257/aer.p20161027} with the goal of creating better public services \cite{mcbride_how_2019}. 

We provide a case study of using open data to analyze the fairness of predictive analytics and provides interventions to improve it.  Previous case studies of other domains include predictive policing~\cite{ensign2018runaway} and child maltreatment~\cite{chouldechova2018case}.

\section{Background}
\label{sec:background}

The starting point for our work is a project done by \citeauthor{schenkjr15a} for the City of Chicago~\cite{schenkjr15a}.
They sought a predictive model to prioritize the scheduling of routine food inspections conducted by sanitarians from the Chicago Department of Public Health (CDPH). Rather than relying solely on manual scheduling of food inspections by CDPH, in 2015 the Chicago Department of Innovation and Technology and data scientists from the Civic Consulting Alliance created a machine learning model to aid in this scheduling process.

The model specifically looks at the scheduling of routine food inspections which are conducted once or twice a year at each establishment independent of any consumer complaints.  The dataset used to train and test the model consists of information from ~18,000 inspections over 4 years with the training set from September 2011 to April 2014 and the testing set from September 2014 to October 2014. The dataset is derived from several datasets from the Chicago Open source portal\footnote{\url{https://data.cityofchicago.org/}} including those about crime rates, sanitation, weather, and food inspections. The dataset also includes information about the sanitarians who conducted the inspections. In order to protect the individual identity and the privacy of the sanitarians, they are grouped into six sanitarian clusters based on their \textit{critical violation rate}, which is the percentage of inspections they conduct that result in critical violations. The clusters, which are used as features in the model, were named after the lines of Chicago rail transit system: Purple, Blue, Orange, Green, Yellow, and Brown. 
From this dataset, they train a logistic regression model using features including the sanitarian cluster conducting the inspection, past establishment violation records, and surrounding environment data (crime rates, cleanliness, temperature) to predict the likelihood of the inspection resulting in a critical violation. 

The trained model is then used to prioritize food inspections. In particular, it outputs a risk score for each inspection where a higher score signifies a higher risk of finding a critical violation. These risk scores are used to prioritize inspections: the highest risk score should be inspected first and the remainder inspected in decreasing order of the risk scores. The input features to the model and its coefficients are detailed in \Cref{tab:city-model-coeff}. 

Evaluating the model requires making an assumption on how it would be used. 
\citeauthor{schenkjr15a}~\cite{schenkjr15a} reassigned the dates of the inspections in the test set based on their predicted risk score, while preserving the number of inspections performed each day and the identity of the sanitarian performing each inspection. (See \Cref{app:reordering} for an illustrated example.)
As their main goal was to ensure that the inspections which resulted in critical violations were inspected as rapidly as possible, they evaluated each schedule (the original and the one created by the model) by the average time required to detect critical violations.  This was calculated by taking the mean of the number of days between date the inspection was scheduled and the first day of the test set window (i.e. September 1, 2014) for those inspections that resulted in a critical violation. They found that, on average, the schedule based on their model detected critical violations 7-days faster than the original schedule when applied to the two-month test set.
The dataset and code used in this analysis are available in the project's repository.\footnote{\url{https://github.com/Chicago/food-inspections-evaluation}}

Kannan, Shapiro, and Bilgic~\cite{kannan2019hindsight} provided an independent analysis of the model results and identified several issues with the model and its analysis.  
As we are primarily interested in the fairness of the \citeauthor{schenkjr15a} approach, many of their findings such as questioning whether some model assumptions hold in practice or arguing that a richer feature set should have been used are not immediately relevant for us (although we revisit some of them in \cref{sec:discussion}).  However, one of their findings turns out to be particularly relevant for our analysis.

In particular, they argued that using the sanitarian clusters as a predictive feature unfairly changes the prediction risk for the establishment. Since the clusters were created by grouping sanitarians with similar violation rates, it was likely that establishments set to be inspected by sanitarian clusters with a high propensity to find violations were much more likely to have a high risk score for a potential violation, regardless of the other attributes.
They view this outcome as being unfair to the restaurant. Although we share this concern, our primary focus is on fairness {\em to customers} residing close to the establishment.  Nevertheless, we show that this differentiated behavior of sanitarians and its use by the model has important consequences for our fairness concerns as well.

\section{Fairness of the Schenk Jr.~et al.~Model}
\label{sec:naive-analysis}

In \cref{sec:background}, we described the logistic regression model used by \citeauthor{schenkjr15a}\cite{schenkjr15a} to schedule inspections for the City of Chicago such that critical violations are found early. In this section, we focus on the fairness of the way the model distributes these public health benefits among the residents of Chicago. To quantify fairness, we borrow from the existing definitions of the fairness literature and adapt them to the problem of food inspections. The first definition we focus on is Demographic Parity (DP) or Statistical Parity, defined as a classifier having equal positive predicted rates for advantaged ($A=1$) and disadvantaged ($A=0$) groups~\cite{calders_building_2009}. Given a prediction $\hat{Y}$, demographic parity is satisfied if
\begin{align}
    \label{eq:demographic-parity}
    P(\hat{Y}=1|A=0) = P(\hat{Y}=1|A=1)\;.
\end{align}
We are interested in achieving similar amounts of time taken to complete food inspections across groups of interest. Since we consider multiple groups, $A$ is categorical with values $\{a^0, \ldots, a^{n-1}\}$, where $n$ is the total number of groups. Let $T$ represent the random variable for the time to complete a random food inspection. Our interpretation of DP is
\begin{align}
    \label{eq:food-inspection-time}
    \E[T|A=a^i] &= \E[T|A=a^{i+1}] \qquad \text{s.t.} \; 0 \leq i < n\;.
\end{align}
\Cref{eq:food-inspection-time} intuitively states that if, on average, the time to conduct the inspections belonging to different groups is equal, then the schedule is fair.

Another widely applicable definition of fairness is Equal Opportunity (EOpp), which is defined as the classifier having equal true positive rates for advantaged and disadvantaged groups~\cite{hardt2016equality}. Formally,
\begin{align}
    \label{eq:equal-opportunity}
    P(\hat{Y}=1|Y=1, A=0) = P(\hat{Y}=1|Y=1, A=1)\;.
\end{align}
Our interpretation of equal opportunity requires having similar times to detect critical violations in food inspections across all groups. This can be written as
\begin{align}
    \label{eq:food-violation-time}
    \E[T|A=a^i, Y=1] &= \E[T|A=a^{i+1}, Y=1] \qquad \text{s.t.} \; 0 \leq i < n\;.
\end{align}
\Cref{eq:food-violation-time} states a schedule is fair if the inspections {\em where a critical violation was found $(Y=1)$}, on average, took equal amounts of time to be detected across the groups. 
These interpretations of DP and EOpp are consistent with work on extending these concepts beyond simple classification settings~\cite{blandin2021fairness}.

Throughout the rest of the paper, we consider an early detection of a critical violation to be an advantage to the people living in the restaurant's neighborhood, as it prevents them from any potential food-borne illness stemming from unsafe conditions. Thus, our primary interest is in applying these definitions to groups consisting of restaurants located in a particular region of the city.
While we consider both the notions of fairness in \Cref{eq:food-inspection-time,eq:food-violation-time}, since the overall goal of the system is to improve the detection of critical violations we put more emphasis on the second.

We largely follow the methodology used by the City of Chicago as described in \cref{sec:background} and fully detailed in \cite{schenkjr15a}. However, to improve the robustness of the results we perform a cross-validated evaluation, rather than the evaluation on the last 60-days performed by \citeauthor{schenkjr15a}.
The dataset contains 19 non-overlapping periods spanning 60 days from the first inspection date till the last. Out of these 19 evaluations periods, we exclude three evaluation periods that did not contain inspections for all 60 days. We report the mean and standard error across the remaining 16 evaluation periods of two metrics: the time to detect a critical violation and the time to conduct inspections.

\subsection{Fairness along geographic lines}
\label{subsec:fairness-geo-race}

\begin{figure}[h!]
    \centering
    \includegraphics[width=0.9\linewidth]{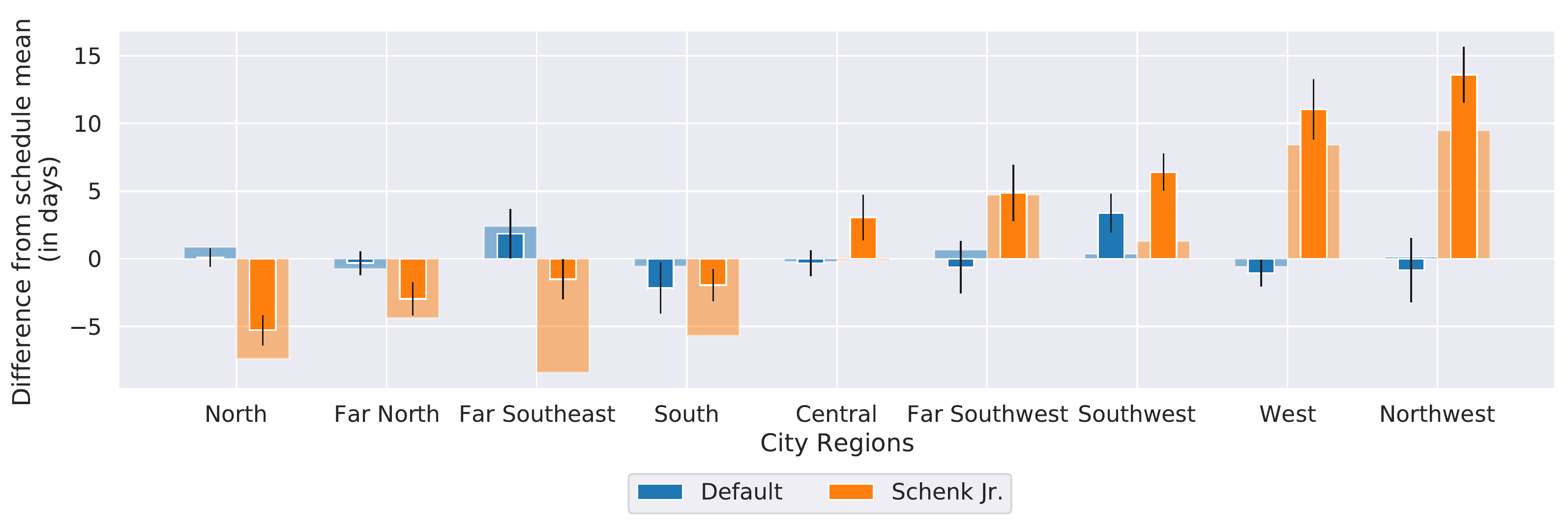}
    \caption{The figure illustrates the difference between the mean time to detect a critical violation in a particular region and the overall mean time for that schedule (EOpp) using the narrower, solid-colored bars. The wider, light colored bars represent the difference in time to conduct inspections regardless of whether a violation was found(DP). The labels represent the major regions of Chicago. Error bars indicate the standard error of EOpp from 16-fold cross validation. (The error bars for DP are similar and omitted for legibility.)
    }
    \label{fig:fairness-by-sides}
\end{figure}

For our fairness analysis, we examine the effects of the model on different regions of the city, which are colloquially known as "sides". We explore how the residents of the different regions are affected by using the model to schedule the food inspections. Using the ZIP codes of the inspected restaurants in the dataset, we match their location to the nine sides\footnote{For a map, see \url{https://en.wikipedia.org/wiki/Chicago\#/media/File:Chicago_community_areas_map.svg}. Accessed 06/12/2021} of the city.

\Cref{fig:fairness-by-sides} shows a disaggregated view of the difference between the average time taken to detect critical violations in a specific region and the overall average for that schedule (which corresponds to the extent to which EOpp is violated in that region, \cref{eq:food-violation-time}) using solid colors.  It also shows the difference between the average time taken to conduct inspections (regardless of whether a critical violation is found) in a region and the overall average for that schedule.  This corresponds to DP (\cref{eq:food-inspection-time}) and is shown using light colors.
The two schedules we consider here are: (a) ``Default Schedule" (blue bars), which is the schedule of inspections that the sanitarians originally followed as they conducted inspections and (b) ``Schenk Jr. Schedule" (orange bars), which is the schedule obtained using the \citeauthor{schenkjr15a} model risk scores and reordering the inspections based on the scores such that inspections with a high risk score (i.e. high predicted likelihood of being a critical violation) are conducted earlier.
The bars indicating negative values signify that the detection times are quicker than the schedule mean (the group is better off than average) and the positive values show that the detection times are slower than the schedule mean (the group is worse off than average).

Considering the Default schedule, we observe all of the regions have detection times close to the schedule mean, consistent with a random schedule being perfectly fair. On the other hand, four out of nine sides have quicker detection times than the average under the Schenk Jr. schedule. For the remaining five that are worse off, two sides receive a far greater delay (at least 10 days) in detecting critical violations.
The trends for the inspection times are similar and suggest that a large part of the gain in critical violation detection for the advantaged regions under the Schenk Jr. schedule comes from inspections in those regions being moved earlier as a whole rather than specifically the inspections most likely to find critical violations. The breakdown of the detection times by region underscores the disparate outcome the Schenk Jr. schedule would have on food inspections in different regions of the city. If used, an individual's place of residence can determine if they have an expedited or delayed routine inspection of food establishments in their neighborhood, which in turn impacts their likelihood of being subjected to a food-borne illness. 

While our primary focus is on this unfairness along geographic lines, we have also explored approaches to assess the fairness of the schedule along racial and economic lines.  As our analysis finds only small effects for these groupings, which are more complex to link to a restaurant than geography, we defer the analysis to \Cref{app:demographic} and \ref{app:economic}.

\subsection{Exploring the cause of unfairness}
\label{subsec:cause-unfairness}

An examination of the coefficients of the \citeauthor{schenkjr15a} model (available in  \Cref{app:model}) shows that the sanitarian conducting the food inspection is a key feature.
As the dataset clusters multiple sanitarians together to protect their identity, we examine sanitarian behavior at the cluster level.
We first inspect the critical violation rate for each of the sanitarian clusters. The critical violation rate is computed as the ratio of the number of inspections that resulted in a critical violation and the total number of inspections conducted by the sanitarians belonging to that cluster.
The critical violation rates for the sanitarians in each of the clusters vary widely, as shown in \Cref{tab:violation-rates}. The Purple sanitarian cluster has the highest rate of citing the restaurants with a critical violation with 41\%. On the other hand, the Brown sanitarian cluster has the lowest critical violation rate of 2.5\%. 
Through personal communication with the authors of \cite{schenkjr15a}, we learned that the sanitarians are grouped into six clusters purely based on their critical violation rate.  
This variation in critical violation rate across sanitarians has at least two possible causes: different strictness among sanitarians and different characteristics of the restaurants inspected.
In \Cref{app:paired}, we analyze restaurants which were inspected by two or more distinct sanitarian clusters and confirm that this difference is driven by the sanitarians rather than properties of the population of restaurants a sanitarian cluster inspects.

To further explore the effects of sanitarian critical violation rate on the unfair outcome for Chicago residents, we plot the location of the inspections conducted by different
sanitarians on a map of Chicago using the 
latitudes and longitudes
\begin{wrapfigure}{l}{0.5\textwidth}
    \centering
    \includegraphics[width=0.49\textwidth]{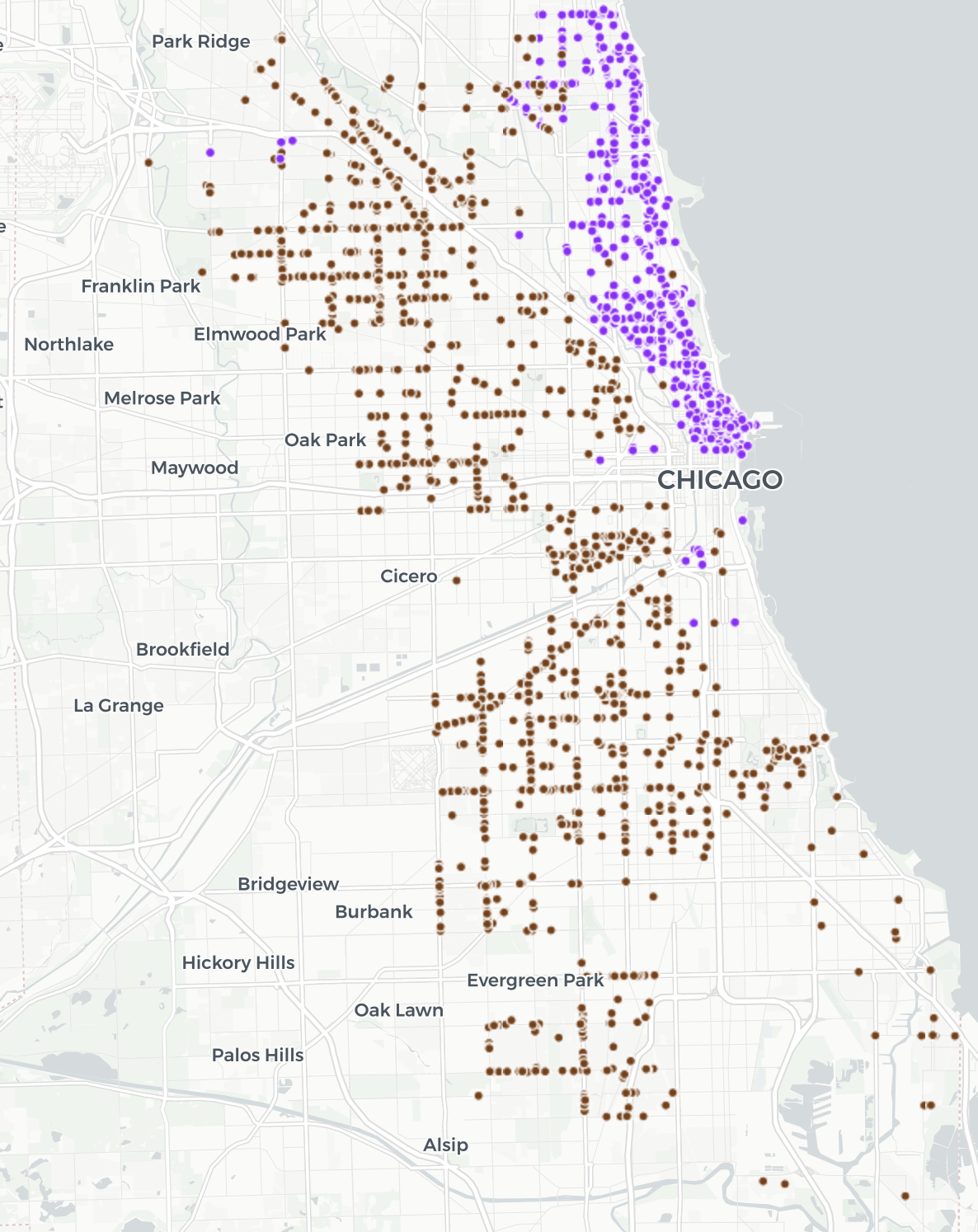}
    \captionof{figure}{Map of Chicago with purple dots representing the food inspections done by Purple cluster sanitarians, and brown dots representing those done by Brown cluster sanitarians. Purple cluster sanitarians and Brown cluster sanitarians have the highest and the lowest critical violation rate respectively.}
    \label{fig:purple-brown-map}
\end{wrapfigure}
from the dataset.
\Cref{fig:purple-brown-map} shows the
inspections done by the Purple cluster sanitarians and those done by the Brown cluster sanitarians. We are particularly interested in these two clusters because they represent the sanitarians with the highest and the lowest critical violation rates. We observe that the inspections conducted by Purple cluster sanitarians are concentrated in the North and Central parts of the city. In contrast, Brown cluster inspections are scattered around in the Northwest and Southwest parts of the city. Therefore, the residents living in the North and Central parts of the city are more advantaged by having a smaller time to detect a critical violation detection than the residents living in the other parts of the city.
See \Cref{app:geography} for maps for all sanitarian clusters.

Finally, we plot the difference in detection and inspections times from the schedule means broken down by sanitarian clusters (rather than by regions as was done in \Cref{fig:fairness-by-sides}) under the default and Schenk Jr. schedules in \Cref{fig:model-sanitarian-violation}. 
Despite varying violation rates, we observe that under the Default schedule all clusters of sanitarians both detect critical violations (solid blue bars) and conduct inspections (light blue bars) at around the same time on average. 
This shows inspections for different clusters were scheduled at roughly equal times, regardless of their results. On the other hand, under the Schenk Jr. schedule (light orange bars) the inspections are sorted in the order of the critical violation rates (\Cref{tab:violation-rates}). The inspections by the Purple sanitarian cluster are scheduled first, and those by the Brown sanitarian cluster are scheduled last. The average times to detect the critical violations follow a similar trend (dark orange bars).
This provides further evidence that the model effectively schedules inspections done by the sanitarians in the order of their violation rate. 

\begin{table}[h]
    \begin{minipage}[b]{0.35\linewidth}
        \centering
        \begin{tabular}{L{2cm}R{3cm}}
            \hline
            \textbf{Sanitarian Cluster} & \textbf{Critical Violation Rate} \\
            \hline \hline
            Purple  & 40.83\%   \\
            Blue    & 25.53\%   \\
            Orange  & 13.76\%   \\
            Green   & 9.68\%    \\
            Yellow  & 5.94\%    \\
            Brown   & 2.5\%     \\ 
            \hline
        \end{tabular}
        \vspace{0.7em}
        \captionof{table}{Table showing the inspector clusters and their critical violation rate for the inspections conducting during the model evaluation period.}
        \label{tab:violation-rates}
    \end{minipage}
    \hfill
    \begin{minipage}[b]{0.6\linewidth}
        \centering
        \includegraphics[width=\linewidth]{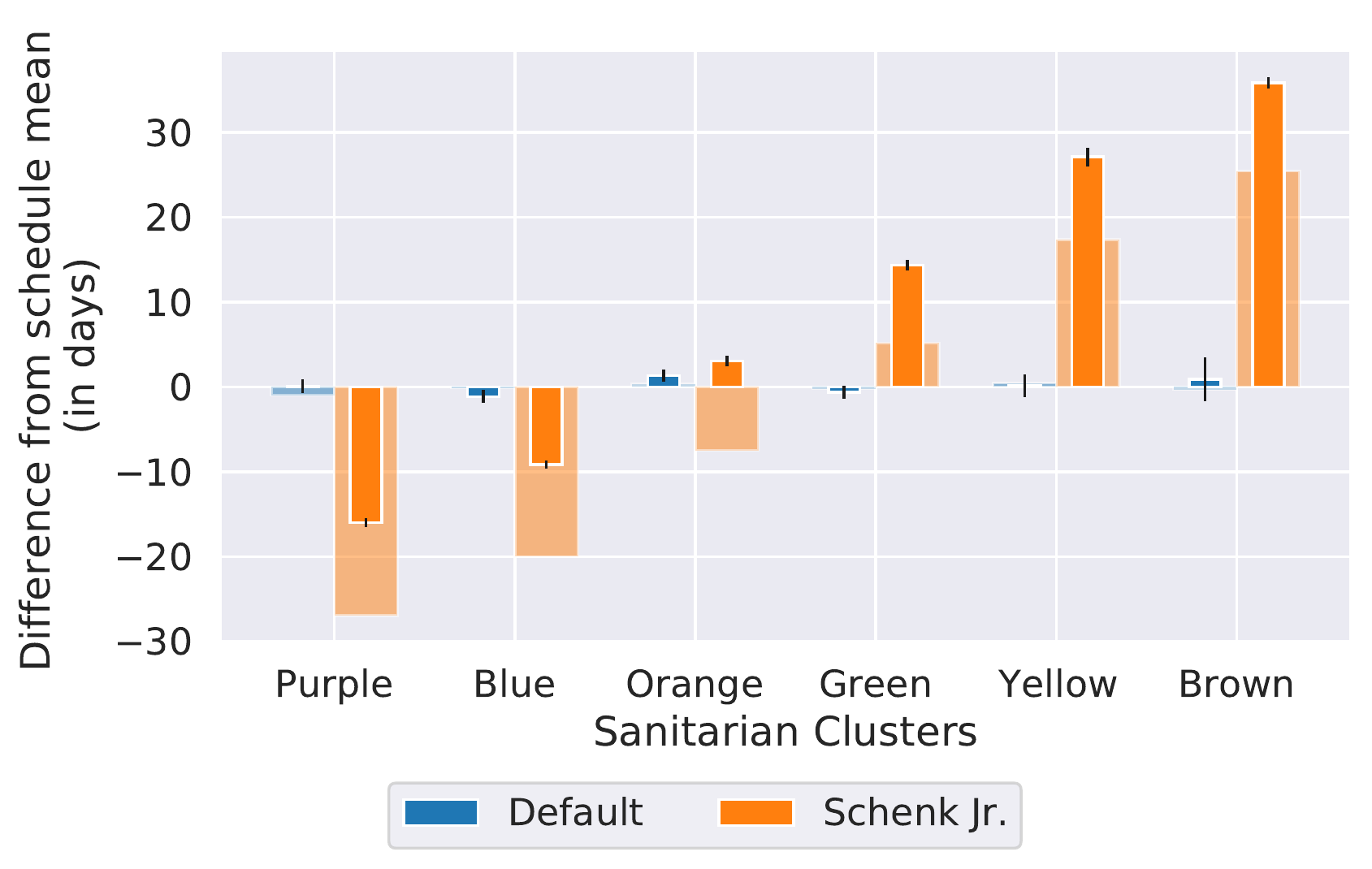}
        \captionof{figure}{The light-colored bars illustrate the time to conduct an inspection (DP) and the solid-colored bars illustrate the time to detect a critical violation (EOpp) relative to the schedule mean. Lower values are better for the cluster.  Error bars indicate the standard error for EOpp.  Those for DP are omitted for legibility.}
        \label{fig:model-sanitarian-violation}
    \end{minipage}
    \vspace{-1em}
\end{table}

To summarize, our analysis suggests that the variation in the violation rates across sanitarian clusters and their significance as features in the \citeauthor{schenkjr15a} model is one of the major causes of geographic unfairness in the resulting schedule.
In the remainder of the paper, we investigate mitigations for both the direct unfairness across sanitarian clusters and the resulting indirect unfairness across regions.
\section{Fairness Through Model Retraining}
\label{sec:model-retraining}

\begin{figure}[h!]
    \begin{minipage}[b]{\textwidth} 
    \centering
    \includegraphics[width=\linewidth]{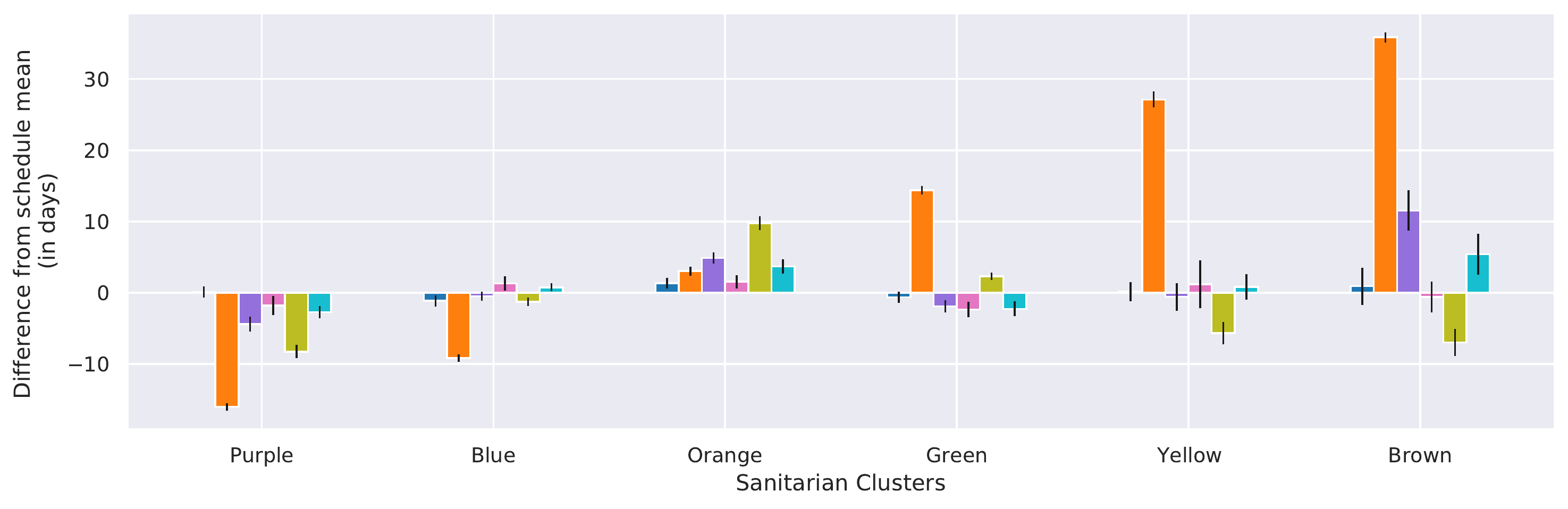}    
    \phantomsubcaption
    \label{subfig:inprocessing-sanitarian}
    \end{minipage}    
    \begin{minipage}[b]{\textwidth} 
    \centering
    \includegraphics[width=\linewidth]{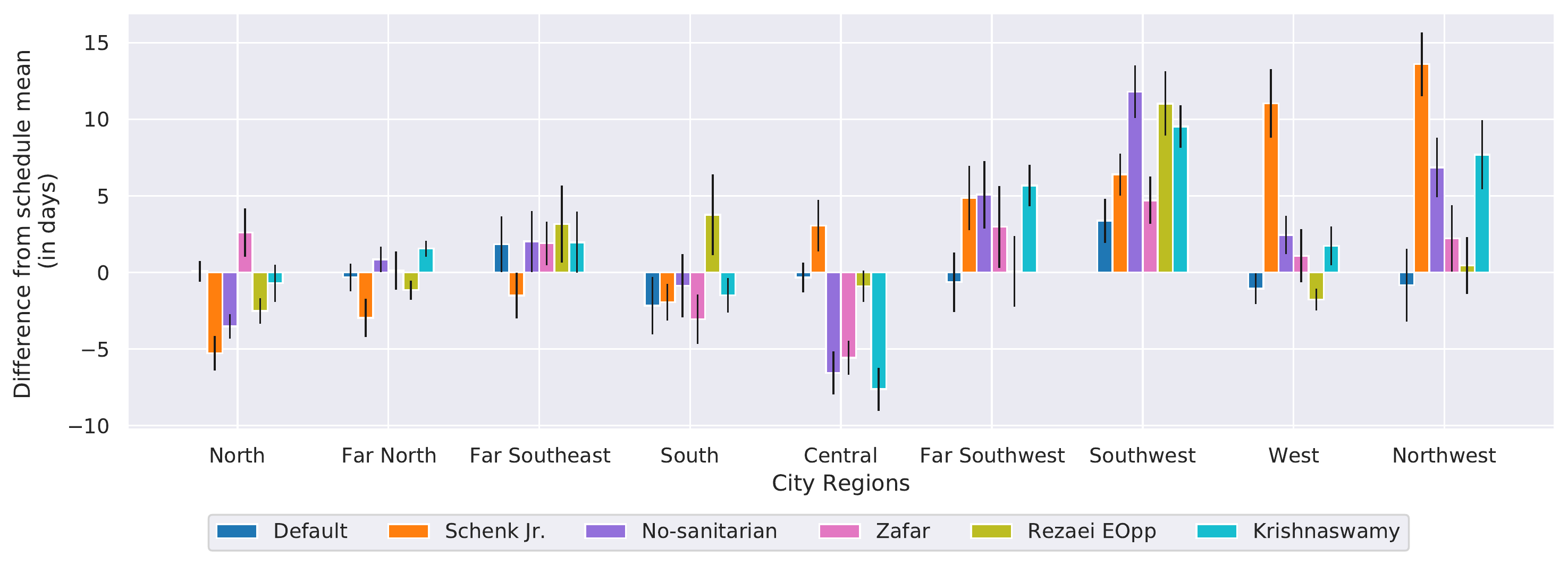}    
    \phantomsubcaption
    \label{subfig:inprocessing-sides}
    \end{minipage}  
    \caption{A disaggregated view of the time to detect a critical violation under four schedules obtained using different model retraining techniques. The bars show the difference in detection times from the schedule mean across sanitarian clusters (\cref{subfig:inprocessing-sanitarian}, top) and geographic groups (\cref{subfig:inprocessing-sides}, bottom) with error bars showing the standard error. (Best viewed in color.)
    }
    \label{fig:sec4-inprocessing}
\end{figure}

In this section, we examine techniques aimed at achieving a fair allocation of food inspection times across sanitarian clusters and city regions by retraining the model in ways designed to result in fairer predictions of critical violations.
We then proceed as in \cref{sec:naive-analysis}.
We use the risk scores from the retrained models to reorder the inspections and measure how fair each approach is by computing the difference from the schedule mean in days.
Our evaluation results preserve the originally assigned sanitarians clusters and the number of inspections done per day.
For brevity, we show only the results for the time to detect a critical violation (EOpp, \cref{eq:food-inspection-time}). Results for DP in \Cref{app:mean-inspection-post} show similar trends.

\subsection{Remove Sanitarians from the Model}
\label{subsec:no-sanitarian-schedule}
For our first approach, we intervene at the \textit{pre-processing} stage.
We train a logistic regression model, the same class of model used by \citeauthor{schenkjr15a}, but do not give the model access to the sanitarian features.
We use the scores from the model to reorder the inspections and call the resulting inspection schedule the ``No-Sanitarian" schedule. 

\Cref{fig:sec4-inprocessing} shows the time to detect a critical violation for the No-sanitarian schedule in purple.
Although the variation under the No-sanitarian schedule reduces in magnitude compared to the Schenk Jr. schedule, the detection times still differ across the sanitarian clusters (\cref{subfig:inprocessing-sanitarian}). 
Inspections done by Purple cluster sanitarians get a higher priority and their mean times are faster than all other sanitarian clusters. Conversely, Brown cluster sanitarians take the most time to detect critical violations. We also see varied detection times across regions (\cref{subfig:inprocessing-sides}). In summary, we observe an improvement over the Schenk Jr. schedule for sanitarian clusters but not a definitive improvement for regions.

Our findings support those from the prior literature~\cite{zemel2013learning,lum2016statistical,barocas2016big,kilbertus2017avoiding} that removing a protected feature, in this case the sanitarian cluster, does not remove bias from the model. We observe that the No-sanitarian schedule follows some of the trends from the Schenk Jr. schedule even though it has no access to the protected attributes. We believe that the correlation of the remaining dataset features with the sanitarian clusters allows the model to continue to discriminate.

\subsection{Fair Regression with Polyvalent Protected Attributes}
Now, we implement the approach proposed by \citeauthor{zafar_fairness_2017} that adds fairness constraints to the logistic regression optimization~\cite{zafar_fairness_2017}.
Their fairness constraints support polyvalent (non-binary) protected features, like the sanitarian clusters in our case. The model enforces a constraint which limits the allowed covariance between the distance to the boundary of the classifier and the protected attributes on the logistic regression loss optimization. Intuitively, this should cause it to avoid the exploitation of correlations we saw with the No-sanitarian schedule.  The allowed covariance is a parameter determining the trade-off between fairness and accuracy. We selected the covariance threshold ($c=0.001$) as the one that produced the fairest outcomes after testing values in ($\{0.0, 10^{-6}, 0.001, 0.01, 0.1\}$).
The resulting scores from the trained model are used to rearrange the inspections and obtain the ``Zafar Schedule".

\Cref{fig:sec4-inprocessing} shows the results for the Zafar schedule in pink. 
The detection times vary less in comparison to the No-sanitarian schedule. In particular, the early detection times for Purple cluster sanitarians and later detection times for the Orange cluster are substantially reduced. 
We see the greatest improvement for the Brown cluster as their detection times are now essentially identical to the overall schedule mean. 
For geographic groups, the regions that were disadvantaged in the Schenk Jr. and No-sanitarian schedules, namely Far Southwest, Southwest, West, Northwest, see considerable improvements. Also, the detection times for the most advantaged regions (North and Far North) are now slightly worse than the schedule mean.
This is consistent with our intuition that the ability of the \citeauthor{zafar_fairness_2017} model to limit the covariance between the decision boundary and the protected attribute should allow it to eliminate the residual effects of the sanitarian features from the dataset and reach a better outcome than the No-sanitarian schedule.

\subsection{Fair Regression with Binary Protected Attributes}
Next, we explore the logistic regression model proposed by \citeauthor{rezaei2020}. It robustly optimizes log loss under an adversarial distribution constrained to lie near the distribution from the data and uses constraints to enforce fairness objectives~\cite{rezaei2020}. Their work focuses on three common fairness objectives: Demographic Parity (DP) ~\cite{calders_building_2009}, Equal Opportunity (EOpp)~\cite{hardt2016equality}, and Equality of Odds~\cite{hardt2016equality}. Since we examine EOpp in this section, we use their model for that objective. 
Since, the \citeauthor{rezaei2020} model requires the protected attributes to be binary, we convert the sanitarian clusters from categorical to binary values by splitting them along their violation rates. We assign the majority protected attribute ($A=1$) to the inspections conducted by Purple, Blue, and Orange cluster sanitarians which a have higher violation rate compared to the rest (\Cref{tab:violation-rates}). Similarly, we assign the remaining inspections done by Green, Yellow and Brown cluster sanitarians to the minority protected attribute ($A=0$). The model allows a regularization parameter $C$, and we select its value ($C=0.5$) that results in the fairest outcome from ($\{0.001, 0.005, 0.01, 0.05, 0.1, 0.2, 0.3, 0.4, 0.5\}$).

We report the results obtained from the \citeauthor{rezaei2020} model under the fairness constraint of EOpp and term them ``Rezaei EOpp Schedule" in \Cref{fig:sec4-inprocessing} in olive. 
We observe that the detection times become less fair compared to the Zafar schedule for the sanitarian clusters although the fairness for city regions is closer.
We believe one of the reasons the \citeauthor{rezaei2020} model does not perform as well as the Zafar model for food inspections is rooted in the loss of information when converting the sanitarian cluster values from categorical to binary sensitive values. For example, nothing prevents the model from delaying Orange cluster inspections to prioritize those of the Purple cluster as the two clusters have been combined.  This emphasizes the importance of developing fair ML models which accept polyvalent protected attributes rather than limiting analysis to the binary case.
Another is that the use of robust optimization means that not only is the model's ability to enforce fairness limited by the need to force it on other nearby models, but that for EOpp in particular there are additional technical complications due to the conditioning on true positives in the definition.

\subsection{Group Proportional Fair Regression}
Finally, we adopt \citeauthor{krishnaswamy_f4a_2021}'s Proportional Fairness classifier~\cite{krishnaswamy_f4a_2021}. Rather than protect specified attributes, they provide guarantees for arbitrary, unknown groups.  This is achieved by training a randomized classifier which guarantees that, for each possible group, the expected utility is in proportion to that of the group's optimal classifier.
The randomized classifier consists of multiple models that are weighted during the training.
To get a single risk score to use when scheduling, we calculate the probability the inspection is predicted as critical (i.e. the sum of weights of classifiers that predict an inspection as critical). We call the schedule obtained from this method the ``Krishnaswamy Schedule", shown in \Cref{fig:sec4-inprocessing} in cyan. The results for the Krishnaswamy schedule are similar to the Zafar schedule and a substantial improvement over the Schenk Jr. Schedule. However, the variation in detection times (for both sanitarian clusters and city sides) is still not close to the near-perfect fairness achieved by the Default schedule. 

To conclude, the approaches we discuss mitigate the sanitarian effect to an extent. However, we believe none of them offer a complete solution as even the fairest (Zafar and Krishnaswamy) still have substantial variation across regions.

\section{Fairness Through Model Usage}
\label{sec:model-usage}

\begin{figure}[h!]
    \begin{minipage}[b]{\textwidth} 
    \centering
    \includegraphics[width=\linewidth]{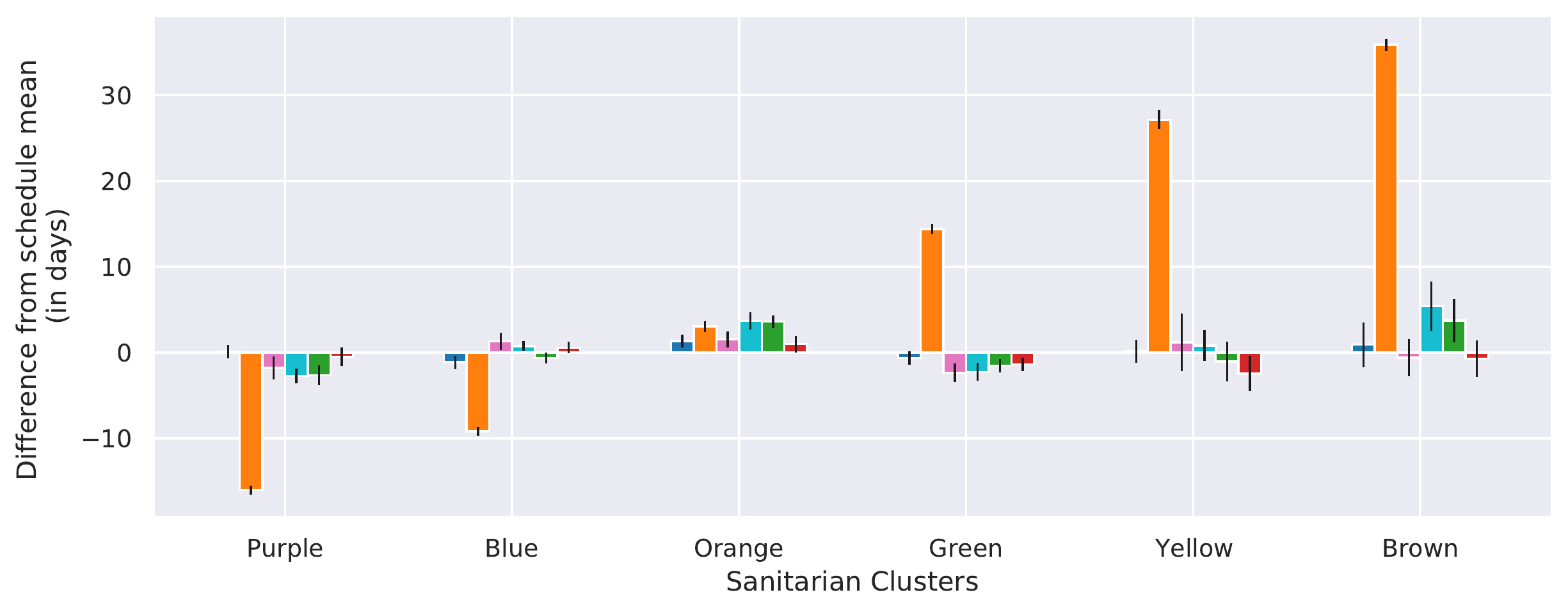} 
    \phantomsubcaption
    \label{subfig:postprocessing-sanitarian}
    \end{minipage}    
    \begin{minipage}[b]{\textwidth} 
    \centering
    \includegraphics[width=\linewidth]{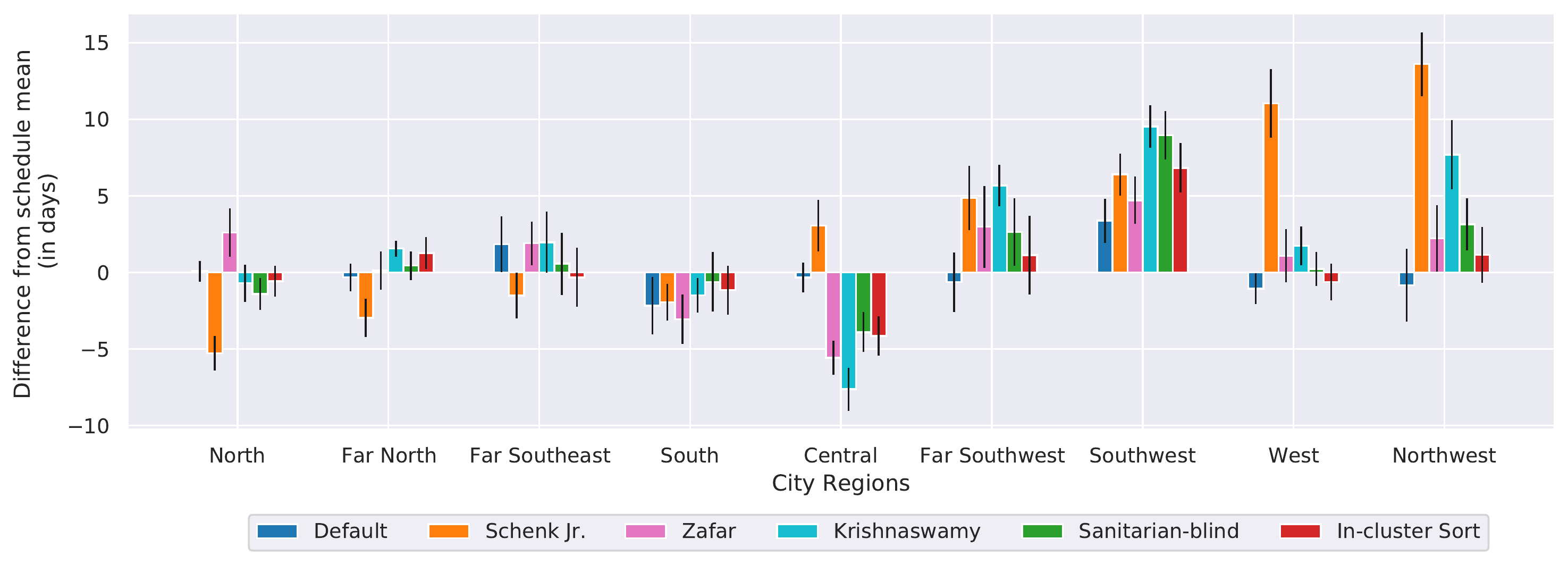}    
    \phantomsubcaption
    \label{subfig:postprocessing-sides}
    \end{minipage}  
    \caption{The times to detect violation under the schedules obtained by post-processing techniques. The bars show the difference from the schedule mean grouped by sanitarian clusters (\cref{subfig:postprocessing-sanitarian}, top) and sides of Chicago (\cref{subfig:postprocessing-sides}, bottom) with error bars giving the standard error. (Best viewed in color.)}
    \label{fig:sec5-inprocessing}
\end{figure}

In this section, we examine two \textit{post-processing} approaches to reduce model disparity. First, we explore suppressing the sanitarian features during model evaluation.
Second, we study the effect of using the model output to schedule the inspections within the sanitarian clusters. As a reminder, we preserve the sanitarian cluster assigned to the inspections in the test set when rescheduling them. 
We present results for EOpp;  similar results for DP are in \Cref{app:mean-inspection-post}

\subsection{Schenk Jr. Schedule with Sanitarians assigned later}
\label{subsec:sanitarian-blind-schedule}
A natural way to use the trained model in practice is by predicting the likelihood of an inspection being a critical violation in the absence of a specific sanitarian and doing those inspections first. 
We do this by keeping the \citeauthor{schenkjr15a} model as-is and setting the sanitarian features to be zero during the evaluation periods. We obtain a new schedule by sorting the inspections by the predicted scores and term it the ``Sanitarian-blind schedule".
This approach is distinct from the No-sanitarian schedule suggested in \Cref{subsec:no-sanitarian-schedule}. That schedule results from eliminating all information about the sanitarian clusters during \textit{training and rescheduling} phases. The Sanitarian-blind schedule does not modify the \citeauthor{schenkjr15a} model but receives no signal related to the sanitarian cluster assignment during \textit{rescheduling}.

In \Cref{fig:sec5-inprocessing}, the detection times for Sanitarian-blind schedule are represented in green.
Broadly, the Sanitarian-blind schedule distributes the detection times among sanitarian clusters similarly to the Krishnaswamy schedule. The Purple sanitarian cluster remains the most advantaged group and the Brown the most disadvantaged. The behavior can be attributed to the fact that while we have blinded the sanitarian features, some of the remaining features correlate with them, as discussed in \Cref{subsec:no-sanitarian-schedule}.
The detection times across regions in \Cref{subfig:postprocessing-sides} reflect an analogous behavior.


\subsection{Schenk Jr. Schedule with In-cluster reordering}
\label{subsec:sanitarian-incluster-schedule}
Another way we could use the model is to first assign each sanitarian a list of restaurant inspections to perform, then use the model to prioritize within each sanitarian's list. The scenario is essentially a ``localized" version of the \citeauthor{schenkjr15a} objective~\cite{schenkjr15a}.
We retain the trained model and its predicted scores using all the features for the evaluation periods. Under all the previous approaches, the inspections can be rearranged based on the predicted score without any constraints. 
For this approach, we consider all the inspections done by each sanitarian cluster separately, sort only those inspections, replace them in the Default schedule, and repeat for each sanitarian cluster. 
In other words, the resulting schedule keeps the number of inspections each sanitarian cluster conducts each day the same as in the Default schedule.
See \Cref{app:reordering} for an illustrated example.
We refer to this schedule as the ``In-cluster Sort Schedule". Unlike the Sanitarian-blind schedule as described in \Cref{subsec:sanitarian-blind-schedule}, the In-cluster Sort schedule does not lose any information during the rescheduling stage and leverages the information gathered from the extra features available.

\Cref{fig:sec5-inprocessing} illustrates the performance of the In-Cluster Sort schedule in red color. The results show the In-Cluster Sort produces a  more equal outcome and the notable differences in the detection times for the Purple and Brown cluster sanitarians from the Krishnaswamy and Sanitarian-blind schedules have become negligible. Correspondingly, \cref{subfig:inprocessing-sides} depicts that the gap in detection times across North and Northwest sides has been bridged as well.  These results are achieved despite the limitations of our data only allowing us to implement this intervention at the level of sanitarian clusters rather than at the intended level of individual sanitarians.

\section{Efficiency and Feasibility}
\label{sec:efficiency-feasibility}

In this section we move beyond fairness alone to consider other important aspects of selecting an approach.
First we examine the trade-off between fairness and efficiency of our approaches. Then we consider how the schedules obtained via different approaches could be used given operational constraints. This raises important questions about the feasibility of the schedules, the choice of right performance metric to evaluate such schedules, and the possibility that some of the efficiency advantages of some methods may be illusory.

\subsection{Fairness and Efficiency Trade-off}
\label{subsec:fairness-efficiency-tradeoff}

We begin by defining our measures of efficiency and fairness. Starting from our definition of Equal Opportunity, we take as our notion of efficiency as the mean time to detect a critical violation:  
\begin{align}
    \mu = \E[T|Y=1] \;.
\end{align}
For fairness, we compute the same metric for each protected group (i.e. sanitarian cluster or region):
\begin{align}
    \mu_i = \E[T|A=a^{i},Y=1]\;.
\end{align}
We then sum the absolute distance of each of the $n$ groups from the overall mean and use this as our fairness metric:
\begin{align}
    d = \sum_{i=0}^{n-1} |\mu_i - \mu |\;.
    \label{eq:unfairness}
\end{align}
This approach is similar in spirit to quantifying the extent to which equal opportunity is violated in a classification setting by comparing the difference in the relevant probabilities between groups.\footnote{We do not report results for DP because they are not meaningful.  Since all schedules preserve the number of inspections conducted each day they by definition have the same efficiency in terms of mean time to conduct the inspections.}

\begin{figure}[h!]
    \centering
    \includegraphics[width=\textwidth]{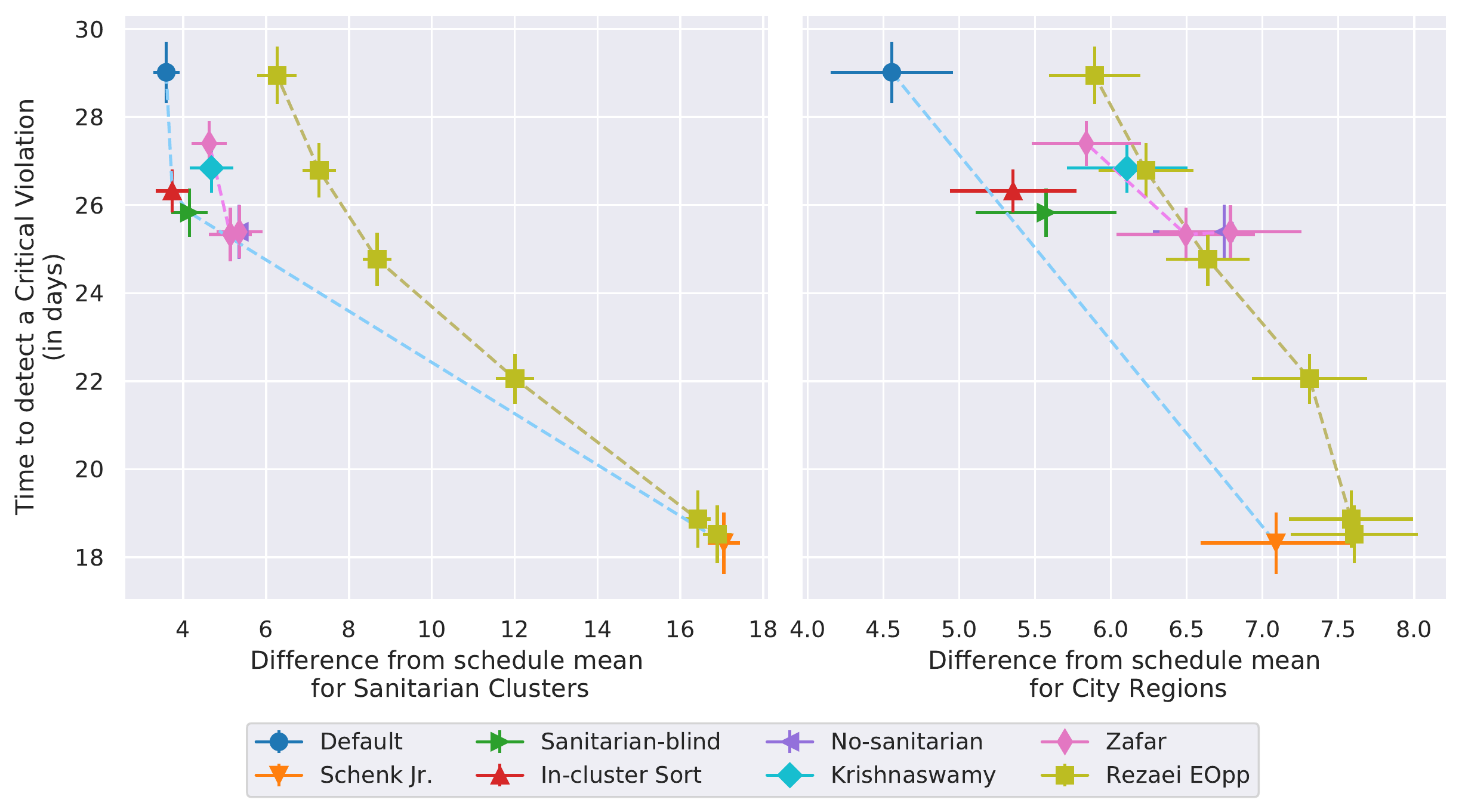}
    \caption{A scatter plot showing the trade-off between the time to detect a violation (EOpp, y-axis) and the fairness which is computed as the average of absolute distance from the mean across all group (x-axis), as defined in \Cref{eq:unfairness}. A lower mean detection time and a lower distance from mean are desirable. Error bars give the standard error from 16 cross-validated runs.}
    \label{fig:efficiency-fairness-tradeoff}
\end{figure}

In \Cref{fig:efficiency-fairness-tradeoff}, we plot the efficiency on the y-axis and fairness on the x-axis. Lower values are better for both.
The Default schedule is the most fair but the least efficient. In contrast, the Schenk Jr. schedule is the most efficient but the least fair to the sanitarian clusters.
The Zafar and Rezaei algorithms have parameters which have the effect of trading off between efficiency and fairness, so for these we plot a range of parameter values ($c=\{0.001, 0.01, 0.1\}$ and $C=\{0.5, 0.2, 0.1, 0.05, 0.01, 0.005\}$ respectively) and illustrate the trade-off curve they enable with dashed lines.
We use a dashed blue line to illustrate the Pareto frontier, the set of schedules that are not dominated in terms of both efficiency and fairness by (a convex combination of) other schedules.
The two model usage approaches lie on or near the Pareto frontier for both sanitarian clusters and regions, indicating they represent trade-offs between efficiency and fairness that may be interesting in practice.  Neither is clearly better than the other. 

Some of the model retraining approaches are near the Pareto frontier for sanitarian clusters, but all are far from it for regions, making their desirability questionable even taking trade-offs between fairness and efficiency into account.
The Zafar schedule varies its efficiency for a relatively small change in fairness. 
As fairness decreases, the Zafar schedule overlaps with the No-sanitarian cluster. This is expected as with higher allowed covariance between decision function and protected attributes the model effectively gets more ability to use the residual sanitarian features.
The Krishnaswamy and Rezaei schedules appear largely dominated, with the exception of Rezaei toward the efficient but unfair part of the Pareto frontier for sanitarian clusters.




\subsection{Operational Constraints}
\label{subsec:city-model-constraints}
From \Cref{sec:background}, we know that the risk scores for \citeauthor{schenkjr15a} model are weighted by the sanitarian cluster. Since the inspections are sorted based on the risk score, the Schenk Jr. schedule assumes that all the inspections are fungible. 

Consider a scenario when all the inspections done by Purple cluster sanitarians are scheduled first. Would it mean that the other cluster sanitarians conduct no inspections during that time and wait for the Purple cluster sanitarians to finish? Do the Purple cluster sanitarians remain idle after conducting their inspections early on? 
Alternatively, if some of the inspections were reassigned to a different sanitarian cluster, do we assume that the result would change or not? 
These questions point to some of the operational constraints encountered in practice and are not accounted for my the methodology introduced by \citeauthor{schenkjr15a}.
A real approach to turning models into schedules needs to be able to account for factors such as limited capacity for a sanitarian to conduct inspections in a day both in terms of the time needed to conduct the inspections themselves and the time needed to travel from inspection to inspection (which depends on where they are located).  Efficiency gains which do not respect these constraints may be illusory.


Such considerations are another advantage of
the post-processing techniques in \Cref{sec:model-usage}.
The Sanitarian-blind schedule works by placing the inspections in an order without needing an assigned sanitarian, allowing later assignment of sanitarians in that respect operational constraints. Likewise, the In-cluster Sort schedule ensures the number of inspections conducted by each cluster each day is reasonable, although it does not account for travel times.\footnote{Ideally we would rearrange inspections at the level of individual sanitarians, but given the limits of the available data we treat all inspections performed by a given cluster as fungible for this purpose.}


\section{Discussion}
\label{sec:discussion}

We have revisited the application of predictive models by the Chicago Department of Public Health to schedule restaurant inspections and performed the first analysis from the perspective of fairness to the population served by the restaurants. We found that the model treats inspections unequally based on the cluster of the sanitarian who conducted the inspection and that there are, as a result, geographic disparities in the benefits of the model.
We examined approaches to using the original model in a fairer way and ways to train the model to achieve fairness and found more success with the former class of approaches.

While our analysis and conclusions are limited to a single data set from the city of Chicago and the particular algorithmic approaches tested, we believe this setting is representative of an important class of problems.  Our communications with experts in food safety suggest that the resource allocation problems and wide differences in violation rates faced by Chicago are common in many jurisdictions.  Beyond food safety, cities conduct a number of other types of inspections including of structural inspections of buildings, fire safety, business licensing, and enforcement of environmental and accessibility regulations.  Thus, we conclude by discussing three broad challenges our results point to for future work.

In contrast to much of the literature that focuses on the fair treatment of individuals, things being inspected are typically businesses or other entities in which many individuals have a stake.  In this work we have taken the simple approach of identifying restaurants with the people who live nearby, but this is certainly a rough approximation at best.  There is a need for better methods to understand who is affected by inspection decisions and how they are affected.
A related problem is understanding and quantifying the effects of inspection scheduling across groups based on race or economic status. The approach we explored (see \cref{app:demographic} and \ref{app:economic}) found limited fairness effects for these groupings but it is unclear whether this is because the algorithms were in fact fair or the approach does a poor job of quantifying the effects so that fairness can be measured.  Beyond simply measuring fairness, developing fair classifications algorithms that can handle the sorts of continuous-valued protected attributes that arise when the data captures the demographic breakdown of a neighborhood or other group of stakeholders is a largely unexplored challenge.

While the goal of the inspections is to protect public health, protection is challenging to measure directly.  Thus we have followed Chicago's approach of using detecting critical violations of the food code as early as possible as a proxy.  The use of such proxies is common, and has caused notable issues in other domains (for example the use of arrests as a proxy for crime~\cite{ensign2018runaway}). The risk of feedback loops has been pointed out in both this and other domains~\cite{kannan2019hindsight,chouldechova2018case}. However, one aspect we wish to stress is that sanitarians have discretion in how they resolve issues they observe, ranging from punishment in the form of critical violations to education and helping restaurant owners correct issues in the course of the inspection.  So a low violation rate is not necessarily indicative of a sanitarian simply missing issues.  Based on our communications with food safety experts, the disparities in violation rates we observe are not unique to Chicago.
Prior work has also found that factors such as the outcome of a previous inspection and the position of an inspection in an inspector’s daily schedule may significantly impact the detection of violations in an inspection~\cite{IbanezMariaR2020HSCB}.
This raises difficult questions about what it means to be fair in such a setting and how to achieve it.  Our approach of reordering within each sanitarian cluster ducks this issue to some extent, assuming what a critical violation ``means'' to a given sanitarian is consistent across time (although even this may not eliminate all issues; see Finding 2 of \cite{kannan2019hindsight}).  However, how can this provide fairness guarantees to individuals?  Are all critical violations really equally bad?  
Given that range of violation rates, it seems likely that some restaurants with no critical violation inspected by Brown cluster sanitarians actually deserve more scrutiny than many restaurants with a critical violation inspected by Purple cluster sanitarians, meaning some part of the increased performance of the original model may be illusory.\footnote{This can also be viewed as an issue of unfairness to restaurants \cite{kannan2019hindsight}.}
What is a better proxy to use for sanitarians who find critical violations only rarely?  Should we be not just reordering inspections but actively shaping which sanitarian performs them in the interest of fairness?

Finally, while the models we use are trained to perform classification, we actually use them for ranking and then those rankings are used for scheduling.  There is room to improve over our approach at all stages of this pipeline.
Would it be good to instead learn a counterfactual ``sanitarian-independent'' violation probability, as is done when predicting clicks in search advertising~\cite{graepel2010web} and has been explored in the literature on causal models in fairness~\cite{kilbertus2017avoiding}?
Rather than trying to achieve fair classification or doing the ranking in ways that address unfairness in the classification, are there better approaches that directly leverage ideas from the literature on fair rankings~\cite{singh2019policy,zehlike2021fairness} or the literature on fair classification in the context of larger systems~\cite{dwork2018fairness}?
We have treated scheduling as a single, static problem, but inspections occur on an ongoing basis.  How should we understand and achieve fairness in the full, dynamic setting?  This last question in particular points to potentially fruitful ways to study this domain in light of the literatures on fairness in reinforcement learning~\cite{jabbari2017fairness,wen2021algorithms} and overall fairness in comparison to local or immediate fairness~\cite{d2020fairness,emelianov2019price,dwork2020individual,liu2018delayed}.

\bibliographystyle{ACM-Reference-Format}
\bibliography{ijcai21}
\clearpage
\appendix
\section{Appendix}

\subsection{City of Chicago Model Parameters and Weights}
\label{app:model}

For completeness we provide a list of the features, descriptions of them, and the coefficient values in the model trained by Schenk Jr.~et al.~in \Cref{tab:city-model-coeff}.  The information in the table is reproduced from~\cite{schenkjr15a}.

\begin{table*}[h]
    \centering
    \begin{tabular}{p{0.3\textwidth}p{0.5\textwidth}r}
    \hline
    \textbf{Variable Name (Literal)}                         & \textbf{Variable Description}                            &  \textbf{Coefficients} \\
    \hline \hline
    \texttt{Inspectorblue}                                   & Indicator variable for Sanitarian Cluster 1              & 0.950\\
    \texttt{Inspectorbrown}                                  & Indicator variable for Sanitarian Cluster 2              & -1.306\\
    \texttt{Inspectorgreen}                                  & Indicator variable for Sanitarian Cluster 3              & -0.244\\
    \texttt{Inspectororange}                                 & Indicator variable for Sanitarian Cluster 4              & 0.202\\
    \texttt{Inspectorpurple}                                 & Indicator variable for Sanitarian Cluster 5              & 1.555\\
    \texttt{Inspectoryellow}                                 & Indicator variable for Sanitarian Cluster 6              & -0.697\\
    \texttt{pastCritical}                                    & Indicates any previous critical violations (last visit)  & 0.302\\
    \texttt{pastSerious}                                     & Indicates any previous serious violations (last visit)   & 0.427\\
    \texttt{timeSinceLast}                                   & Elapsed time since previous inspection                   & 0.097\\
    \texttt{ageAtInspection}                                 & Age of business license at the time of inspection        & -0.164\\
    \texttt{consumption\_on\_premises\_incidental\_activity} & Presence of a license for consumption / incidental activity & 0.411\\
    \texttt{tobacco\_retail\_over\_counter}                  & Presence of an additional license for tobacco sales      & 0.171\\
    \texttt{temperatureMax}                                  & The daily high temperature on the day of inspection      & 0.005\\
    \texttt{heat\_burglary}                                  & Local intensity of recent burglaries                     & 0.002\\
    \texttt{heat\_sanitation}                                & Local intensity of recent sanitation complaints          & 0.002\\
    \texttt{heat\_garbage}                                   & Local intensity of recent garbage cart requests          & -0.004\\
    \hline
    \end{tabular}
    \caption{The features and their corresponding coefficient weights as used by the City of Chicago to forecast restaurant inspections with critical violations.}
    \label{tab:city-model-coeff}
\end{table*}

\subsection{Geographical Distribution of Sanitarian Clusters}
\label{app:geography}
\begin{figure}[h!]
    \centering
    \includegraphics[width=0.8\linewidth]{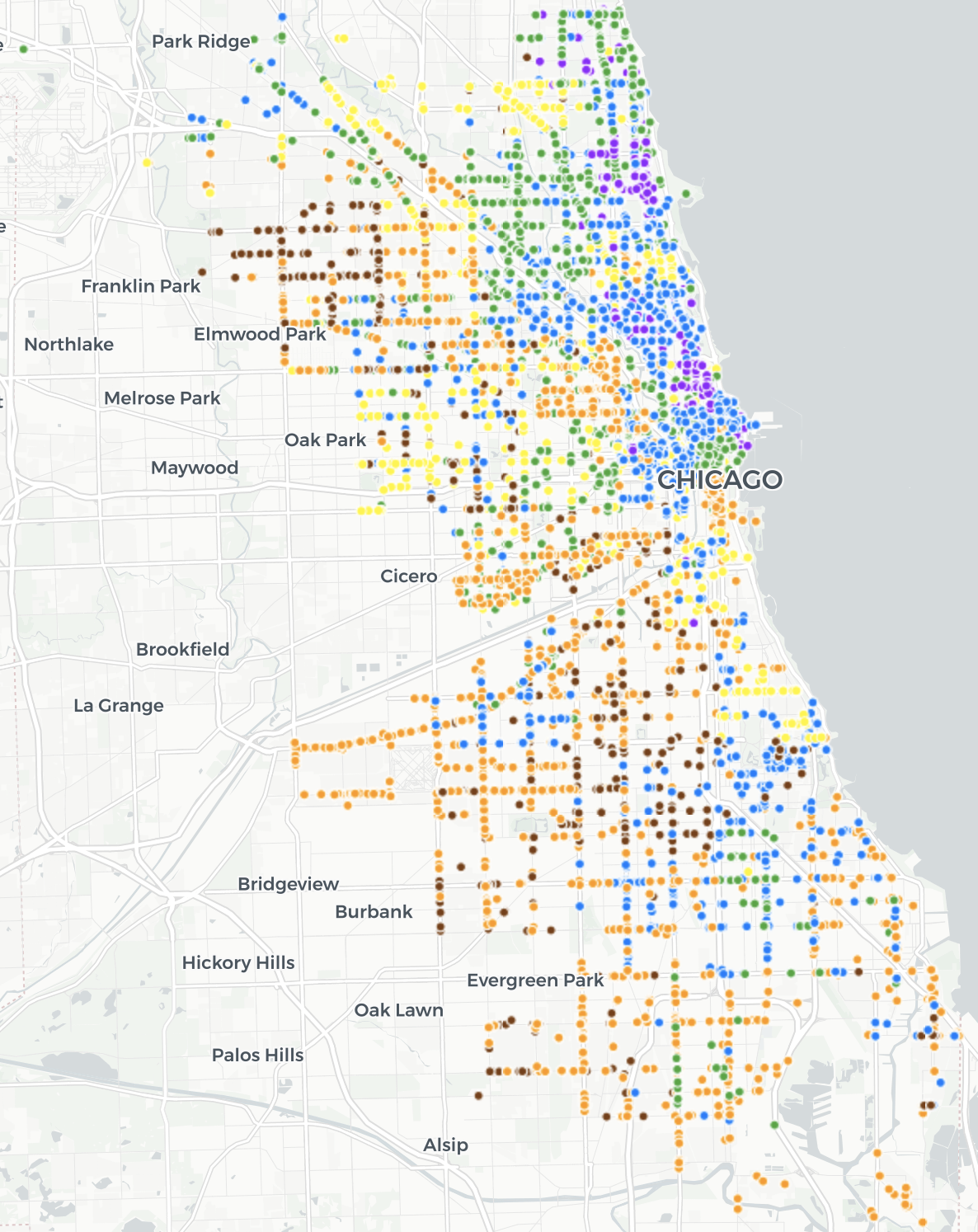}
    \caption{Map of Chicago with dots representing inspections by different sanitarian clusters.} 
    \label{fig:all-maps-in-one}
\end{figure}

\begin{figure*}[h!]
  \centering
\setkeys{Gin}{width=0.3\linewidth}
  \includegraphics{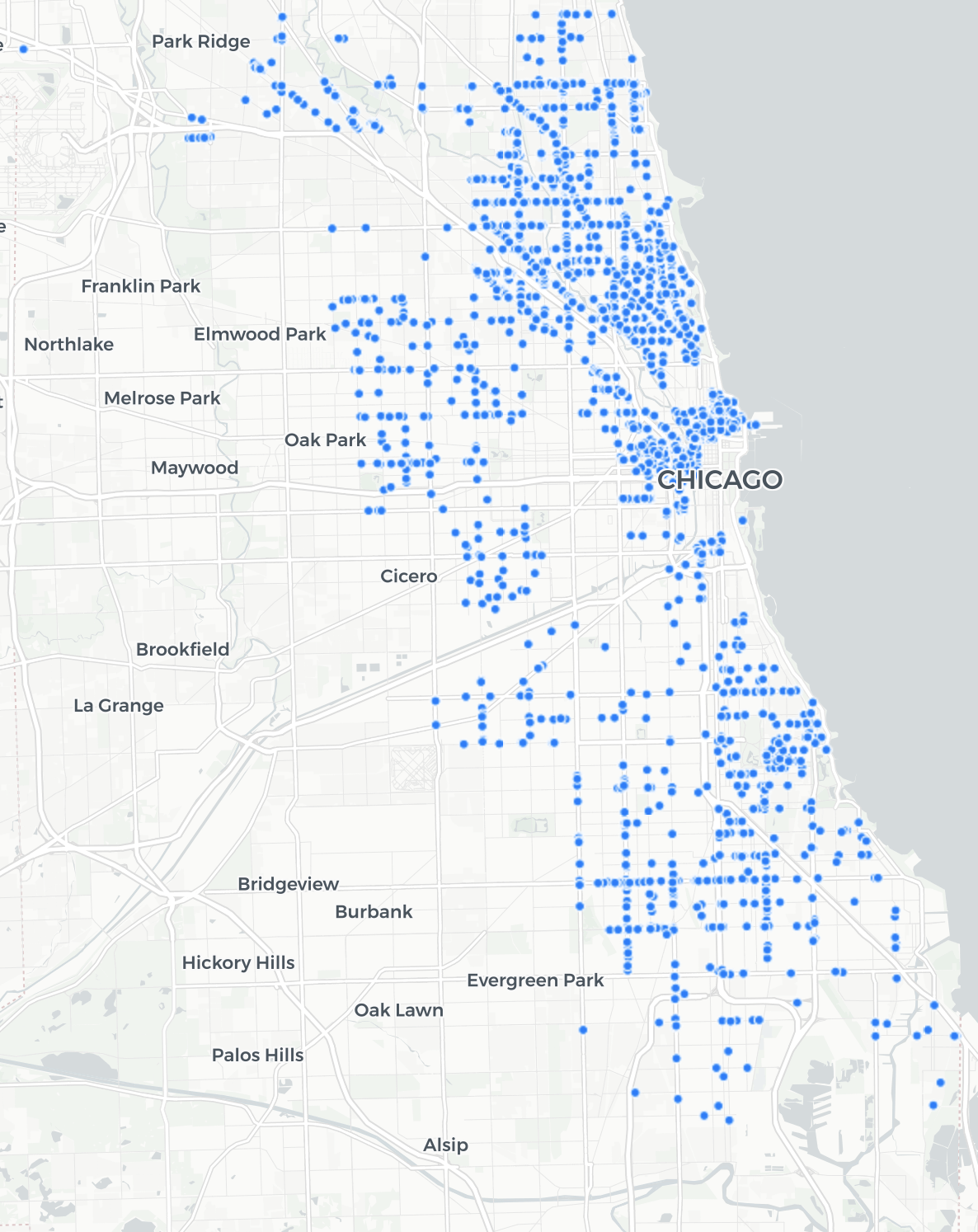}\,%
  \includegraphics{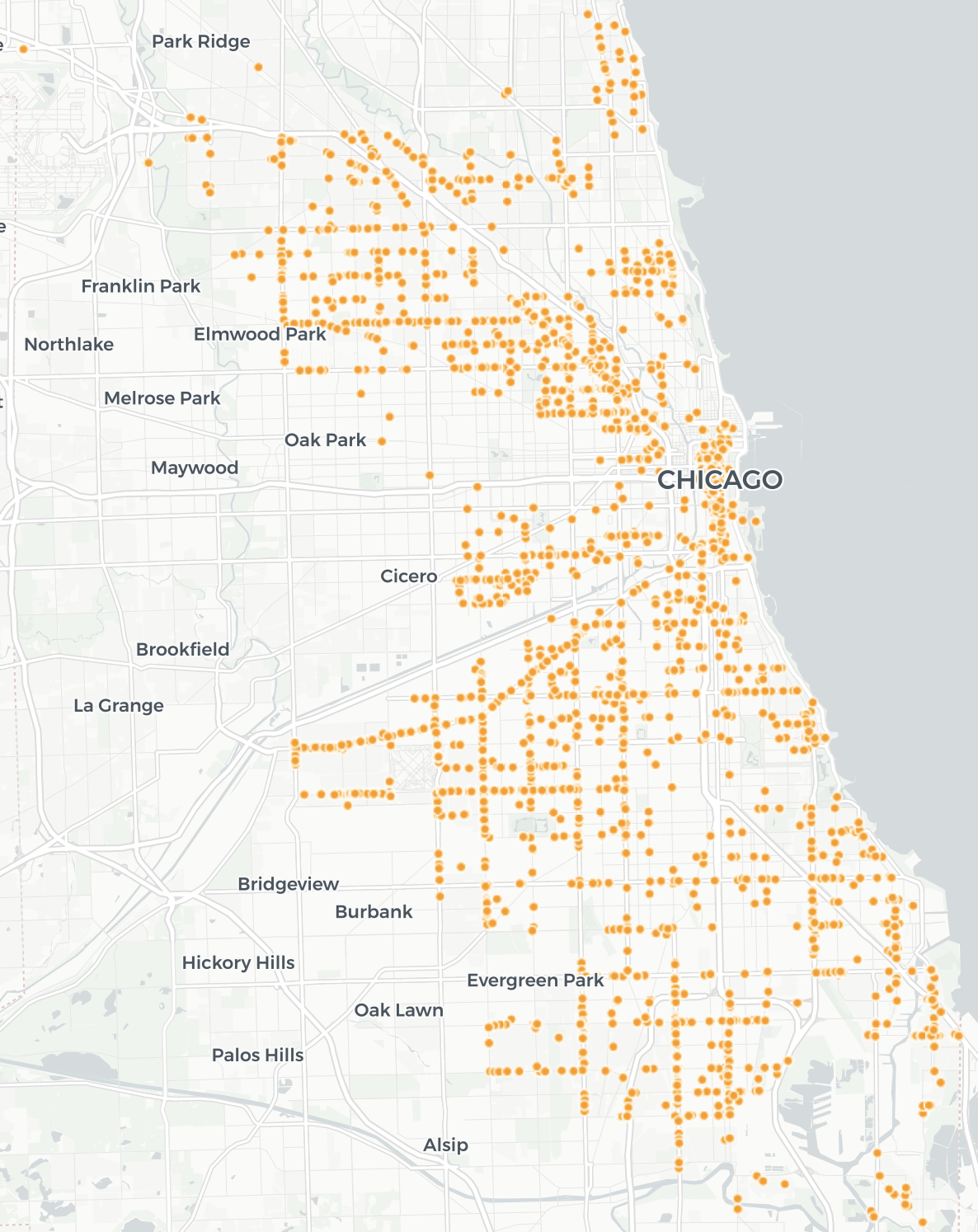}\,%
  \includegraphics{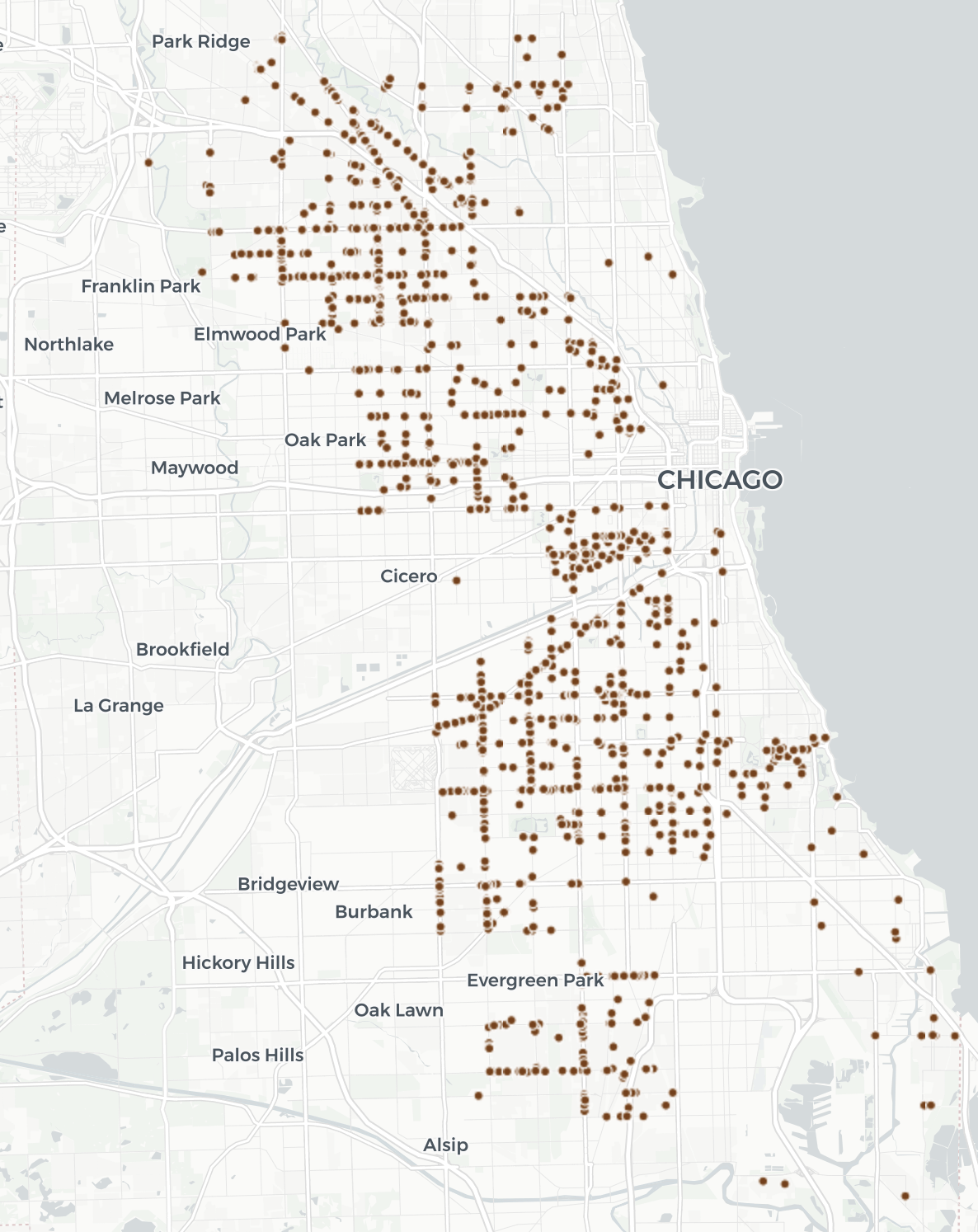}

  \includegraphics{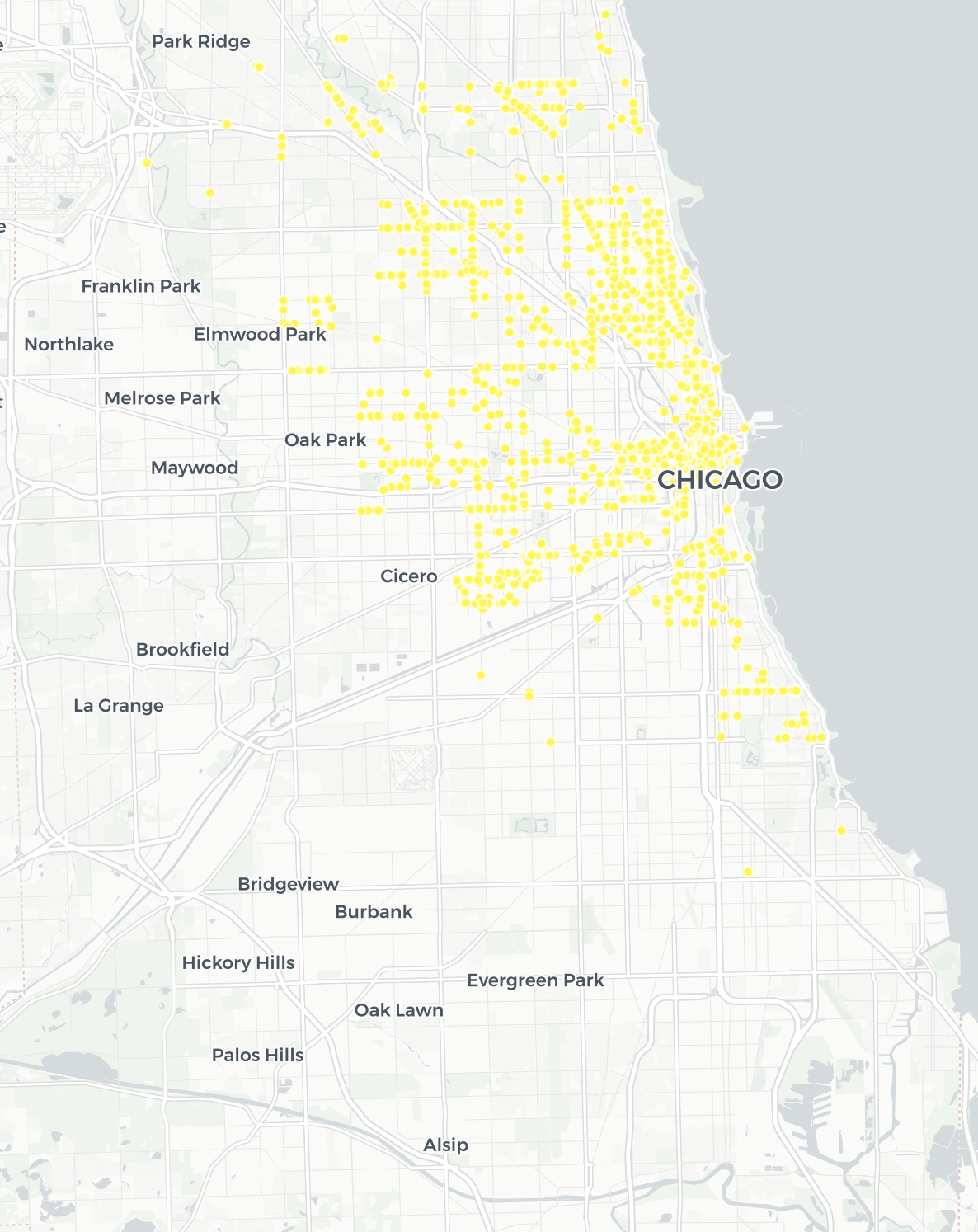}\,%
  \includegraphics{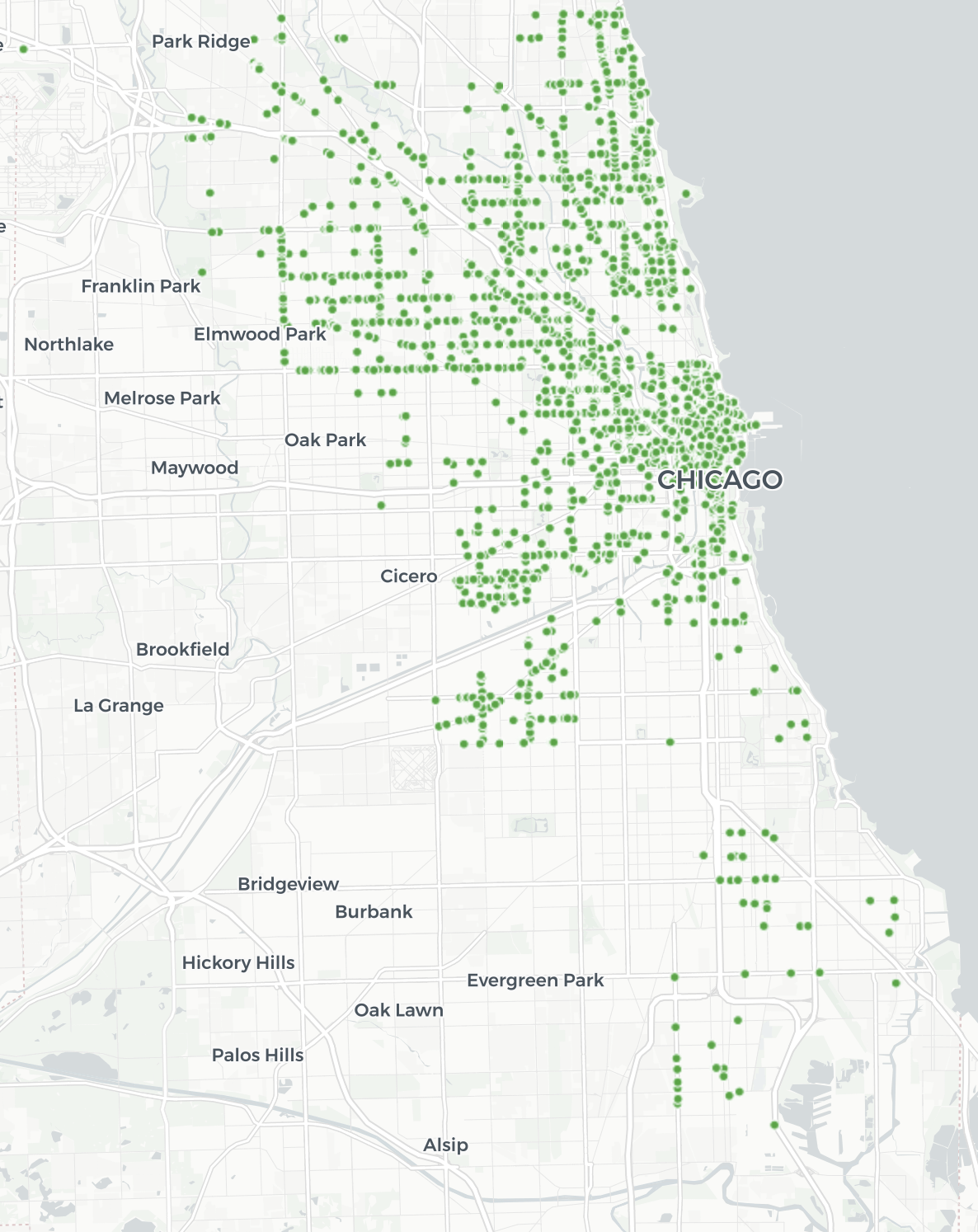}\,%
  \includegraphics{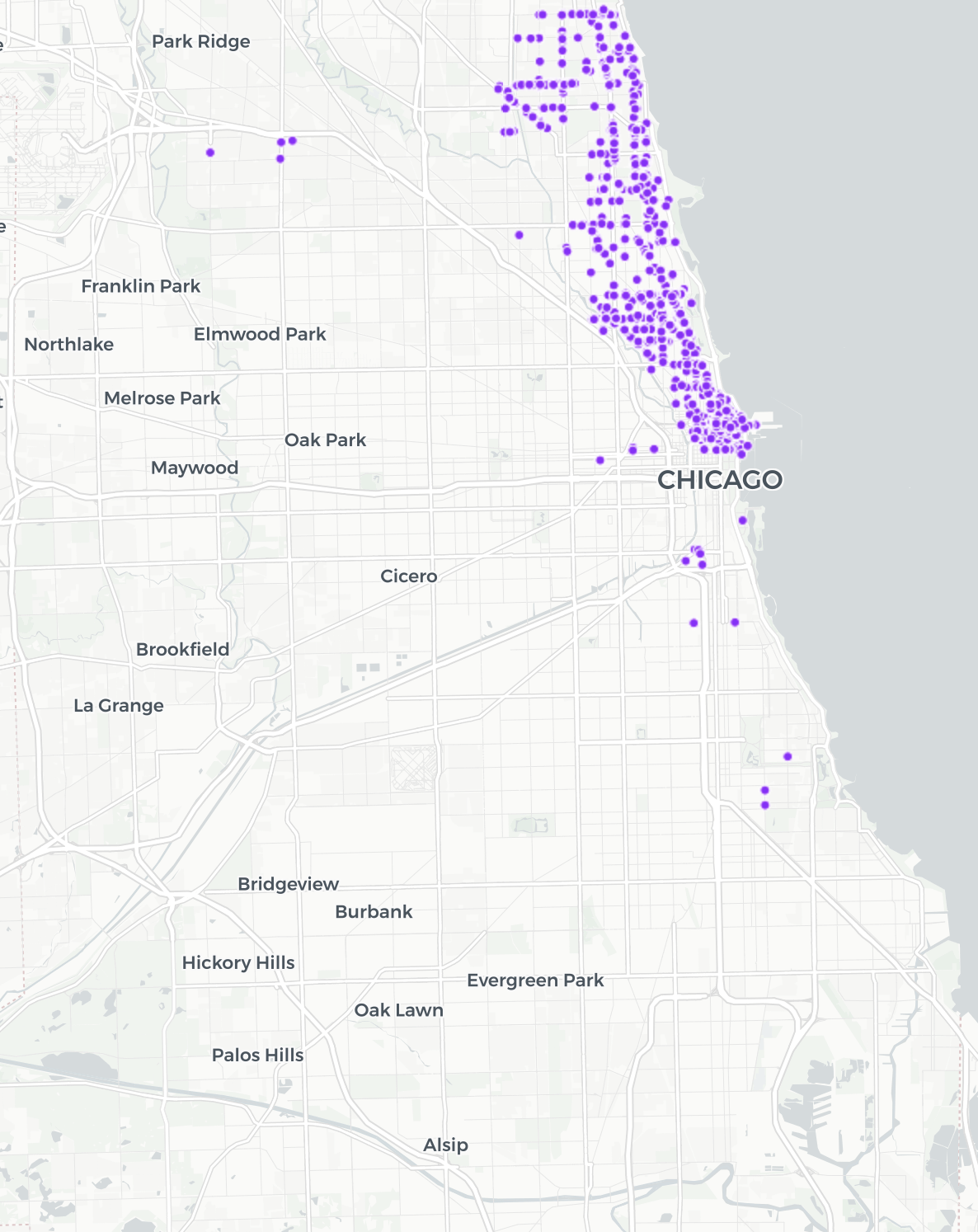}
  \caption{The figure illustrates 6 maps showing the geographical distributions of the sanitarian clusters across the city.}
  \label{fig:all-maps-separate}
\end{figure*}

Figures~\ref{fig:all-maps-in-one} and \ref{fig:all-maps-separate} show each inspection as a dot and the colors represent the sanitarian cluster that completed the inspection. Since the sanitarian clusters were created based on violation rate, these figures also give an idea of where inspections with different rates of violations are concentrated.  
Sanitarians in the purple sanitarian cluster are very concentrated in the north east of Chicago which is consistent with the lower mean critical violation detection times in the Far North Side and North Side shown in \ref{fig:fairness-by-sides}. Others are more dispersed but still show geographic tendencies.
\Cref{tab:san-geo} gives the number of inspections conducted by each cluster in each region, which quantifies this overall effect.


\begin{table*}[h]
    \centering
    \begin{tabular}{*{11}{c}}
    \hline
    \textbf{Cluster}                         & \textbf{CC}                            &  \textbf{Far North} &  \textbf{Far SE}&  \textbf{Far SW}&  \textbf{North} &  \textbf{Northwest} &  \textbf{South}&  \textbf{Southwest}  &  \textbf{West} &  \textbf{Total}   \\
    \hline \hline
    \texttt{Blue}& 344&	799&	161&	176&	1096&	65&	340	&126	&353&	3460\\
    \texttt{Brown}& 9&	265&	22&	343&	51&	371&	51&	485&	418&	2015\\
    \texttt{Green}& 1551&	1223&	50&	112	&658&	379&	125	&133&	727	&4958\\
    \texttt{Orange}& 476&	429&	480	&340&	242&	236&	345	&1050&	494&	4092\\
    \texttt{Purple}& 412&	493&	0&	1&	344&	0&	20&	0&	7&	1277\\
    \texttt{Yellow}& 697&	225&	3&	1&	752&	205&	139&	5&	990&	3017\\
    \texttt{Total}&3489	&3434&	716	&973&	3143&	1256&	1020&	1799&	2989&	18819\\
    \hline
    \end{tabular}
    \caption{Break down of geographical location where clusters have performed inspections on the complete dataset used to train and test the model.}
     \label{tab:san-geo}
\end{table*}

\clearpage

\subsection{Same Establishment Different Sanitarian}
\label{app:paired}

\begin{figure}[h!]
    \centering
    \includegraphics[width=0.6\linewidth]{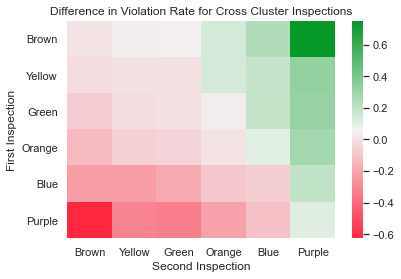}
    \caption{The figure depicts all of the establishments that had two or more inspections in out data set. Each square represents difference between the fraction of inspections in which the row sanitarian did not find a violation and the column sanitarian did from the fraction of inspections in which the row sanitarian found a violation and the column sanitarian did not.} 
    \label{fig:cross-sanitarian}
\end{figure}

As discussed in \cref{subsec:cause-unfairness}, it is a priori unclear whether the differences in the critical violation rates across sanitarian clusters are due to the individual propensities of sanitarians or characteristics of the restaurants those sanitarians inspect.  As evidence that it is primarily due to the sanitarians, we exploit the fact that each establishment is not always inspected by the same sanitarian. In fact, a majority of the establishments have been inspected by sanitarians in different clusters over the 4 year observation period. This provided us with the opportunity look at how different sanitarian clusters assessed the same establishment. We went through each pair of sanitarian clusters and looked for four different instances: both clusters finding a violation, only one of the clusters finding a violation, and neither cluster finding a violation. This provides a number of different comparisons that could be made, and all appear consistent with the sanitarian performing the inspection as the primary cause of the difference. In Figure \ref{fig:cross-sanitarian} we visualize the outcomes of pairs of inspections of the same restaurant.  Specifically, we plot the difference between the fraction of instances where the first inspection did not find a violation but the second one did and the fraction where the first inspection did but the second did not, aggregated by sanitarian cluster.
The top right square's strong green represents brown sanitarians not finding violations when purple sanitarians do in the next inspection substantially more often than brown sanitarians finding violations and purple sanitarians not finding a violation after. The bottom left corner shows the opposite, purple sanitarians not finding violations when brown sanitarians do in the next inspection is much less common than purple sanitarians finding violations and brown sanitarians not finding a violation after. The axes of the plot are ordered by violation rate of sanitarian clusters and the clear color gradient provides strong evidence that the difference is violation rate is due to sanitarian tendencies rather then the nature of restaurants inspected. The entries on the diagonal or not exactly zero because repeat inspections of a restaurant by the same cluster do not always have the same result.
Aggregating across all such paired inspections where at least two clusters conducted an inspection of the restaurant, \Cref{tab:violation-rates-paired} shows that the overall violation rates on such paired inspections are quite close to the overall rates for each cluster (as shown in \Cref{tab:violation-rates}).  This provides additional confidence that the restaurants which were inspected by different clusters are not somehow substantially different from those inspected by a single cluster.

\begin{table}[ht]
    \centering
    \begin{tabular}{lr}
    \hline
    \multicolumn{1}{l}{\textbf{Sanitarian Cluster}} & \multicolumn{1}{l}{\textbf{Critical Violation Rate}} \\ \hline \hline
    Purple                                         & 40.21\%                                              \\
    Blue                                           & 25.21\%                                              \\
    Orange                                         & 15.14\%                                              \\
    Green                                          & 9.15\%                                               \\
    Yellow                                         & 6.68\%                                               \\
    Brown                                          & 2.51\%                                                \\ \hline
    \end{tabular}
    \caption{Table showing the inspector clusters and their critical violation rate for inspections of restaurants also inspected by another sanitarian cluster.}
    \label{tab:violation-rates-paired}
\end{table}

\clearpage

\subsection{Mean Inspection Times For Policies from \cref{sec:model-retraining} and \cref{sec:model-usage}}
\label{app:mean-inspection-post}

\begin{figure}[h!]
    \begin{minipage}[b]{\textwidth} 
    \centering
    \includegraphics[width=\linewidth]{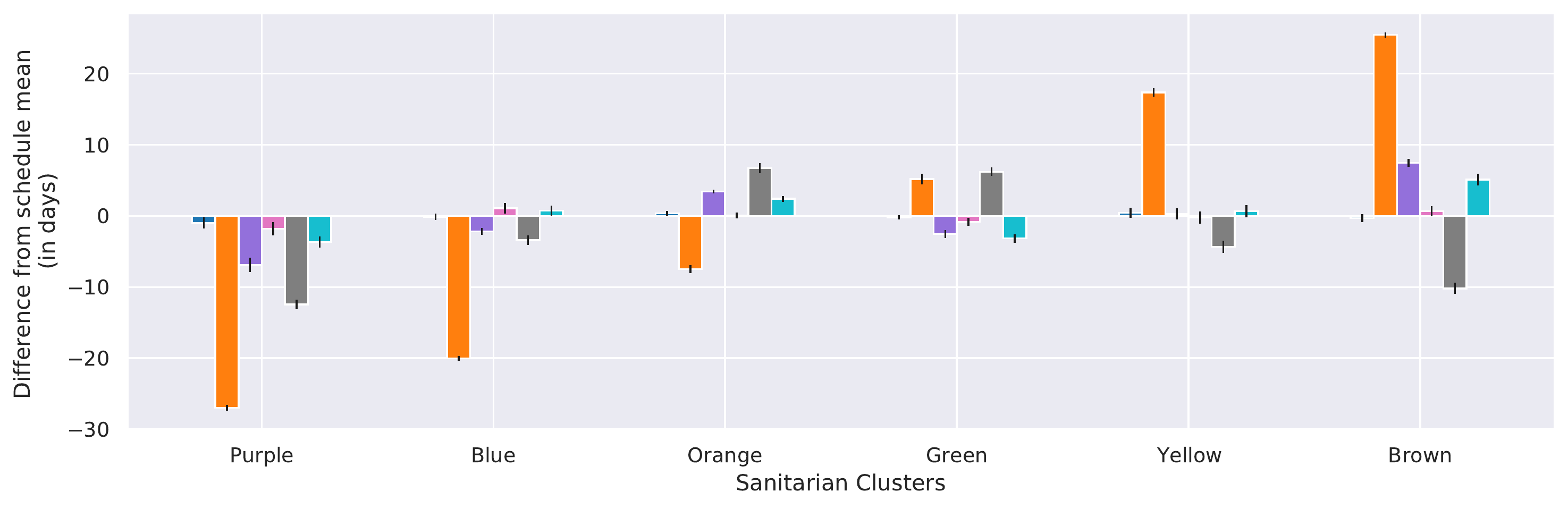}    
    \phantomsubcaption
    \label{subfig:dp-inprocessing-sanitarian}
    \end{minipage}    
    \begin{minipage}[b]{\textwidth} 
    \centering
    \includegraphics[width=\linewidth]{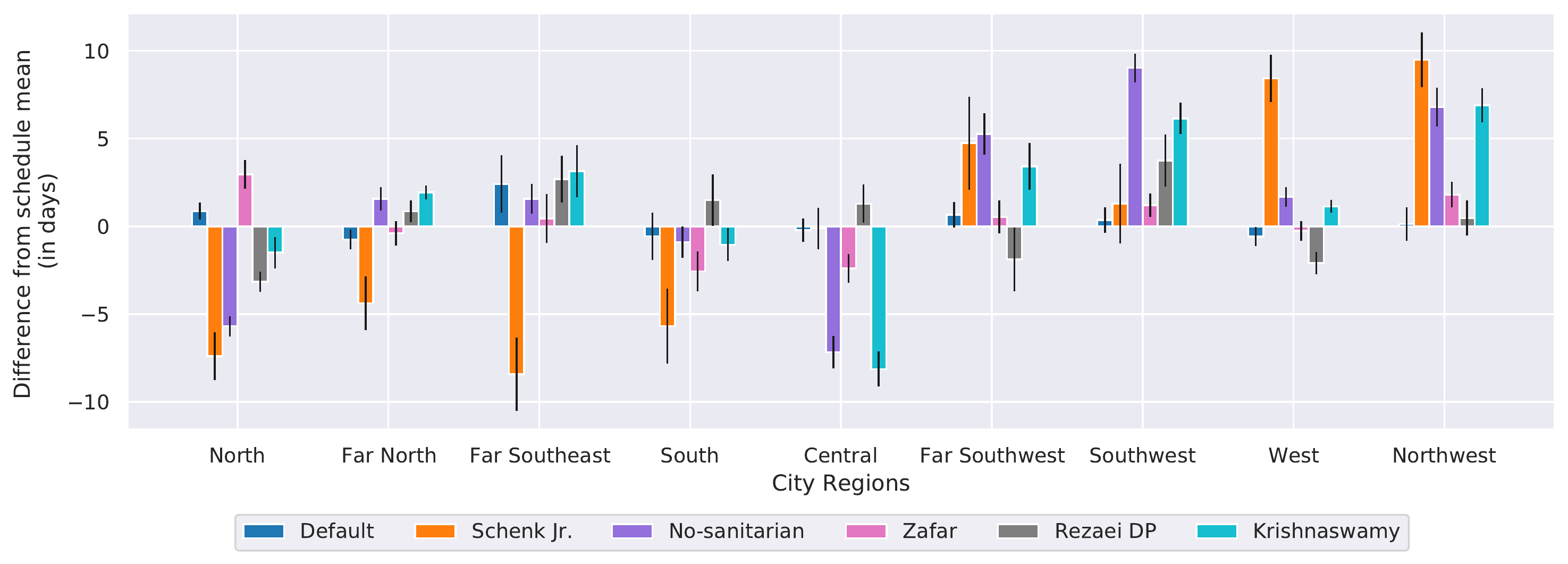}    
    \phantomsubcaption
    \label{subfig:dp-inprocessing-sides}
    \end{minipage}  
    \caption{A disaggregated view of the time to detect a critical violation under four schedules obtained using different model retraining techniques, focusing on demographic parity. The bars show the difference in detection times from the schedule mean across sanitarian clusters (fig. 4a, top) and geographic groups (fig. 4b, bottom) with error bars showing the standard error. (Best viewed in color.)}
    \label{fig:dp-inprocessing-sanitarian}
\end{figure}

\begin{figure}[h!]
    \begin{minipage}[b]{\textwidth} 
    \centering
    \includegraphics[width=\linewidth]{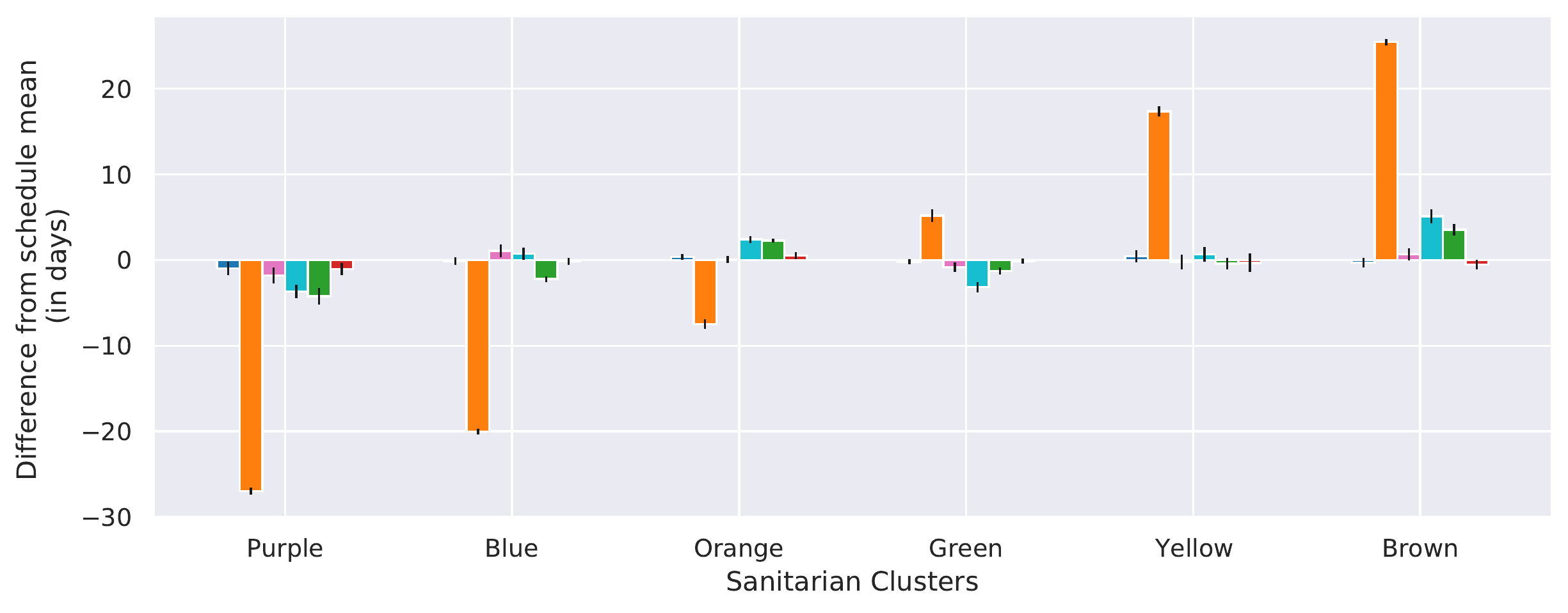}    
    \phantomsubcaption
    \label{subfig:dp-postprocessing-sanitarian}
    \end{minipage}    
    \begin{minipage}[b]{\textwidth} 
    \centering
    \includegraphics[width=\linewidth]{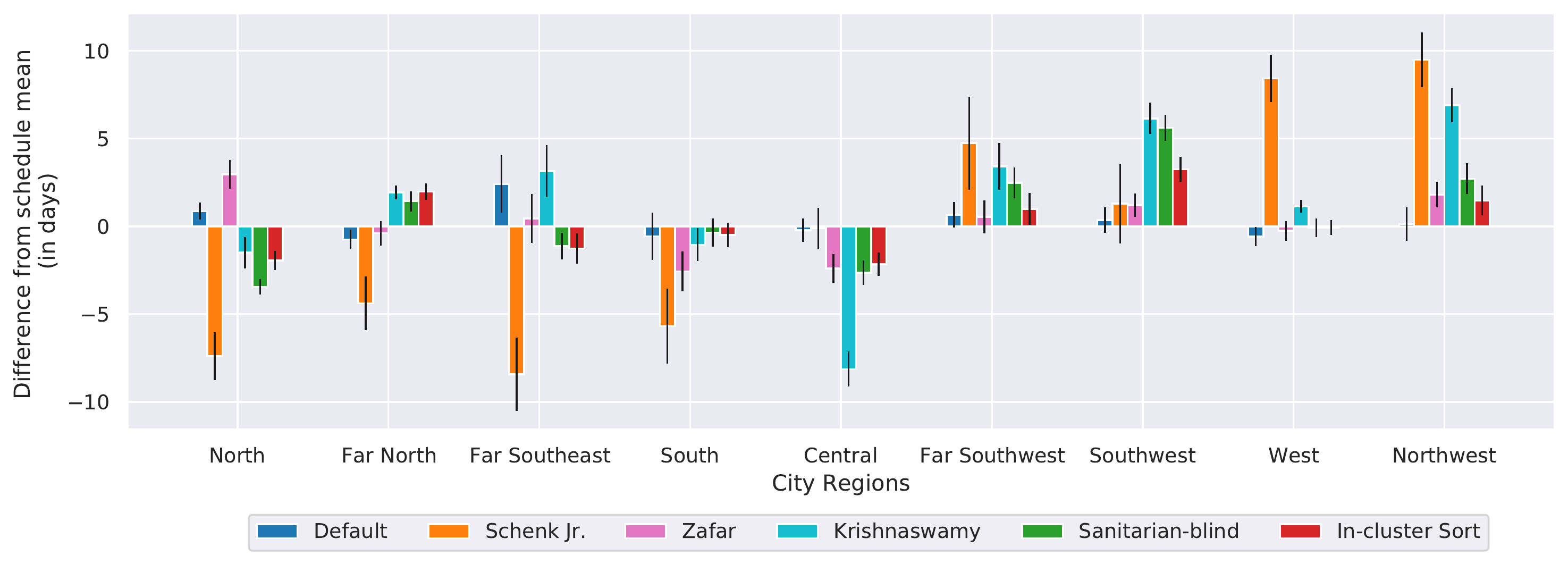}    
    \phantomsubcaption
    \label{subfig:dp-postprocessing-sides}
    \end{minipage}  
    \caption{The times to detect violation under the schedules obtained by post-processing techniques, focusing on demographic parity. The bars show the difference from the schedule mean grouped by sanitarian clusters (top) and sides of Chicago (bottom) with error bars giving the
standard error. (Best viewed in color.)}
    \label{fig:dp-postprocessing-sanitarian}
\end{figure}

In the main text we presented the mean time to detect critical violations in \Cref{fig:sec4-inprocessing} and \Cref{fig:sec5-inprocessing}.  In \Cref{fig:dp-inprocessing-sanitarian} and \Cref{fig:dp-postprocessing-sanitarian}, we present the mean time to conduct the inspections overall.  For the Rezaei schedule we now use the DP version of their algorithm, but otherwise the schedules are unchanged.  A visual comparison of the DP and EOpp figures suggests they are highly correlated and that the improvements or worsening in detection time are driven in large part by simply conducting inspections for that sanitarian cluster as a whole earlier or later respectively.


\clearpage

\subsection{Explaining the Reordering Completed by the Schenk Jr. model}
\label{app:reordering}
\begin{figure}[h!]
    \centering
    \includegraphics[width=0.8\linewidth]{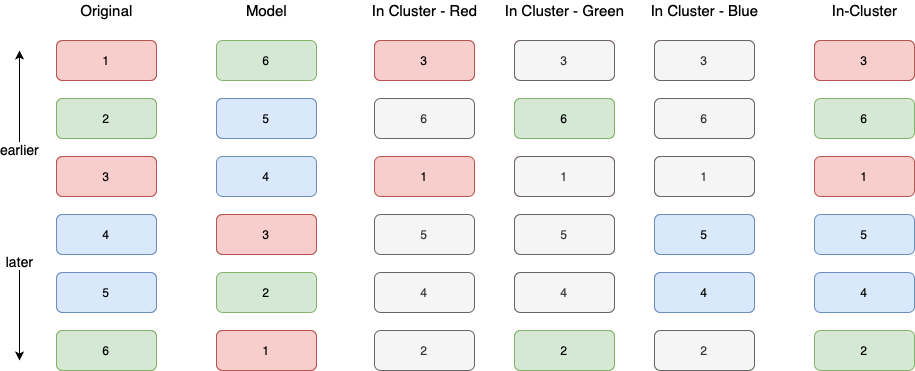}
    \caption{The sanitarian based in-cluster reordering is a schedule where each sanitarian cluster still has inspections at the same time, the establishments are just rearranged.} 
    \label{fig:reordering_graphic}
\end{figure}

Figure \ref{fig:reordering_graphic} provides a visual explanation of both the rescheduling process introduced by Schenk Jr.~et al.~and the in-cluster reordering process used as a post processing technique in \cref{sec:model-usage}. The first column shows a sample initial schedule of six inspections with the colors indicating the cluster of sanitarian assigned to the inspection and the number representing the risk score of the inspection. In the second column, the inspections have been rescheduled in decreasing order of risk score by a model. Note how the identify of the sanitarian conducting the inspection is preserved even as the inspections are reordered. The next three column show a reordering of just the inspections conducted by each sanitarian cluster in isolation.  The In-Cluster schedule in the final column is the combination of these separate reorderings.

\subsection{Fairness Along Demographic Lines}
\label{app:demographic}

In addition to exploring model fairness through the lens of sanitarian cluster and geography, we also examine the fairness of our approaches across demographic groups, specifically race and ethnicity. To do so, we leverage the restaurant location in our dataset and the US Census Bureau data. The US Census publishes the American Community Survey (ACS) every year containing the population estimates using the samples collected from previous years~\cite{census_acs_multiyear}. For our analysis, we use 2014-2018 ACS 5-year estimates for each block group's race and ethnicity composition. The latitude and longitude from the dataset are used to find which block group a restaurant belongs to and gather the race and ethnicity composition for that block group. We get the normalized fractional composition for each block group, and we weight the difference in detection time from schedule mean by the fractional demographic composition for each violation compute the weighted average for each demographic group.

For brevity we focus on four main demographic groups: White, Black, Asian, and Hispanic.\footnote{Other demographic groups such as Native American or population of multiple races are excluded from our analysis because their numbers are low} We also assume that the resident living close to the restaurant all have equal access to its services and experience equal welfare from detecting a critical violation. 

\begin{figure}[h!]
    \begin{minipage}[t]{0.49\textwidth} 
        \centering
        \includegraphics[width=\linewidth]{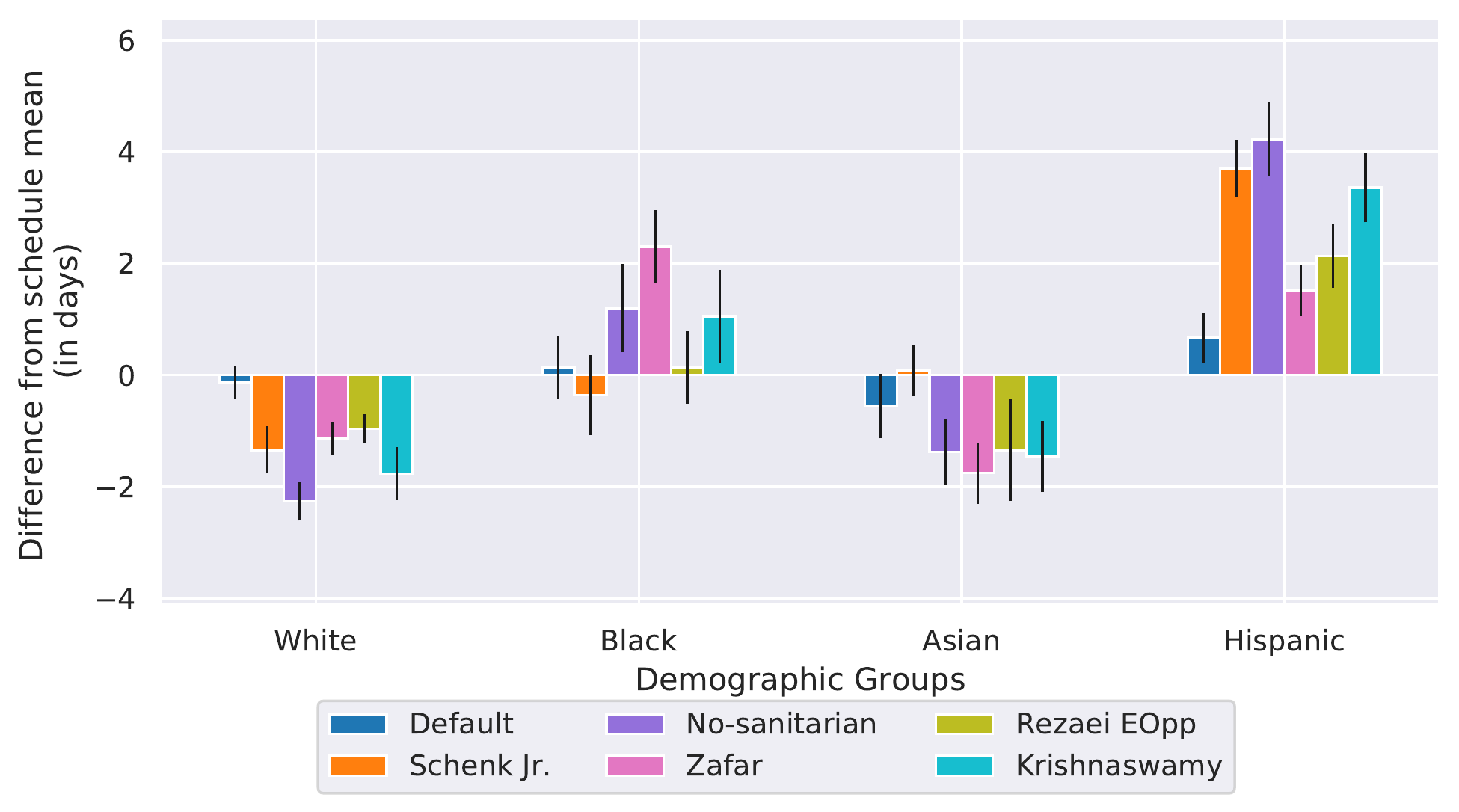}    
        \phantomsubcaption
        \label{subfig:inprocessing-race}
    \end{minipage}    
    \begin{minipage}[t]{0.49\textwidth} 
        \centering
        \includegraphics[width=\linewidth]{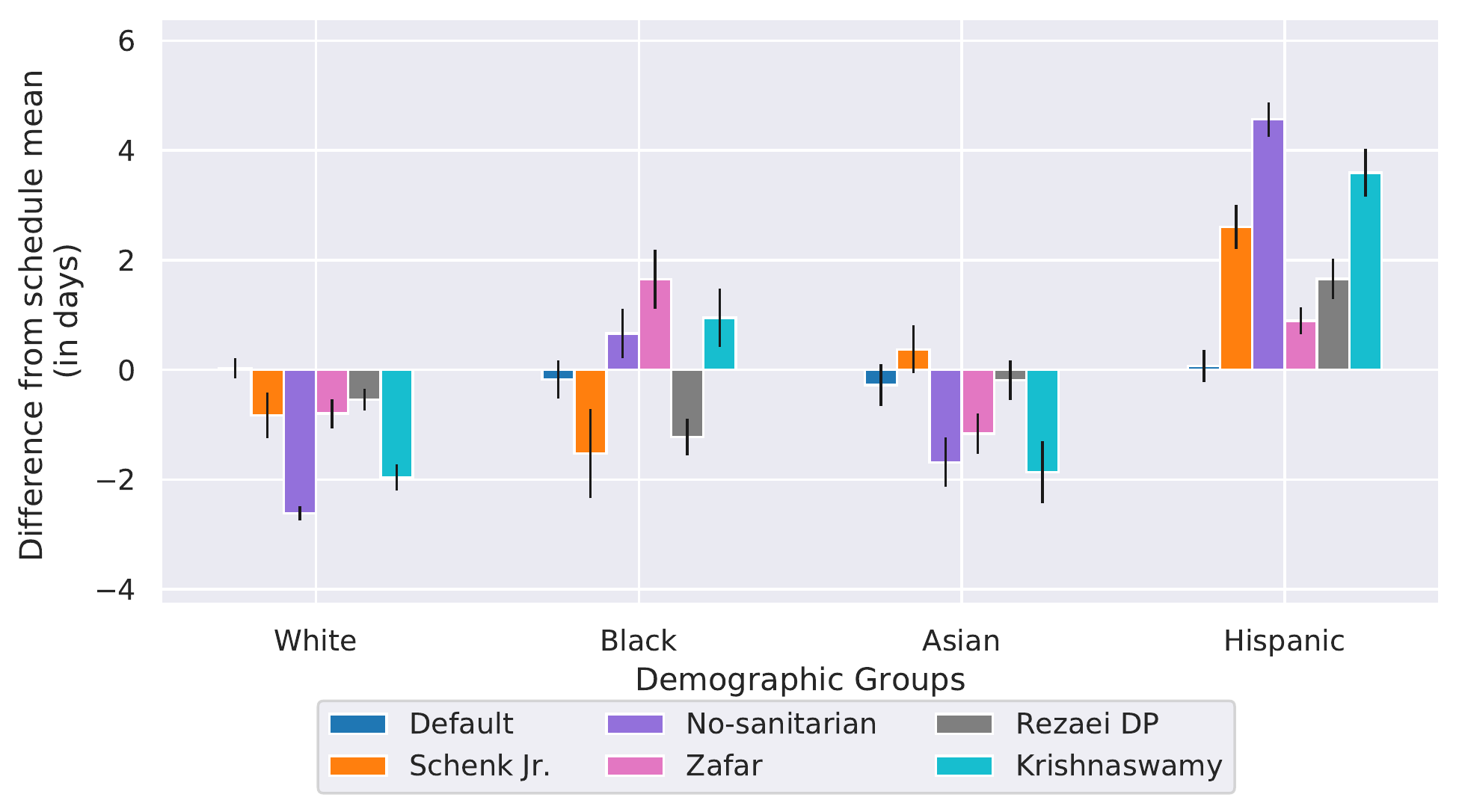}    
        \phantomsubcaption
        \label{subfig:inprocessing-race-dp}
    \end{minipage}
    \caption{The difference in detection (\cref{subfig:inprocessing-race}, left) and inspection (\cref{subfig:inprocessing-race-dp}, right) times for model-retraining approaches, explained in \Cref{sec:model-retraining}. }
    \label{fig:app-inprocessing-race}
\end{figure}

\begin{figure}[h!]
    \begin{minipage}[t]{0.49\textwidth} 
        \centering
        \includegraphics[width=\linewidth]{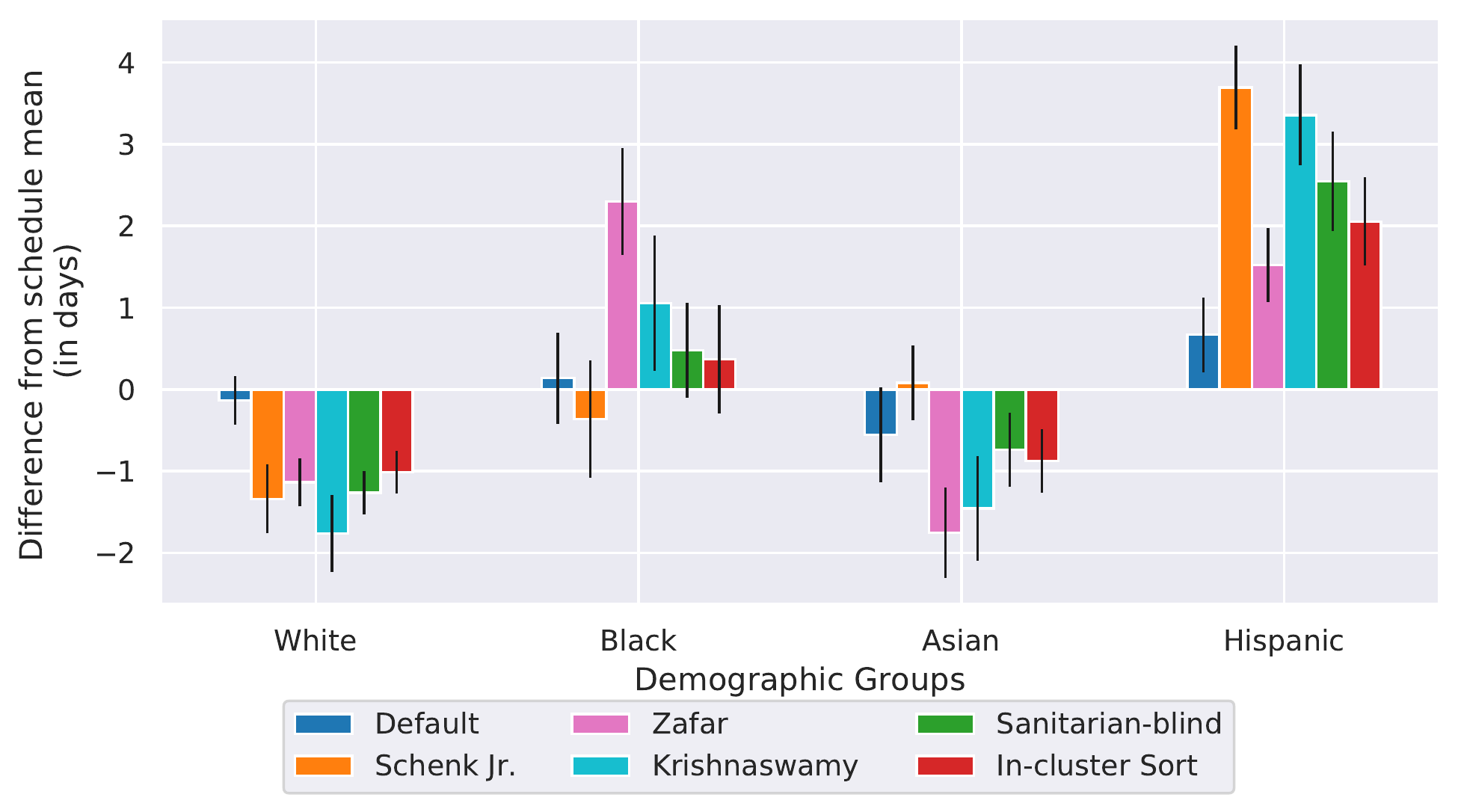}    
        \phantomsubcaption
        \label{subfig:postprocessing-race}
    \end{minipage}    
    \begin{minipage}[t]{0.49\textwidth} 
        \centering
        \includegraphics[width=\linewidth]{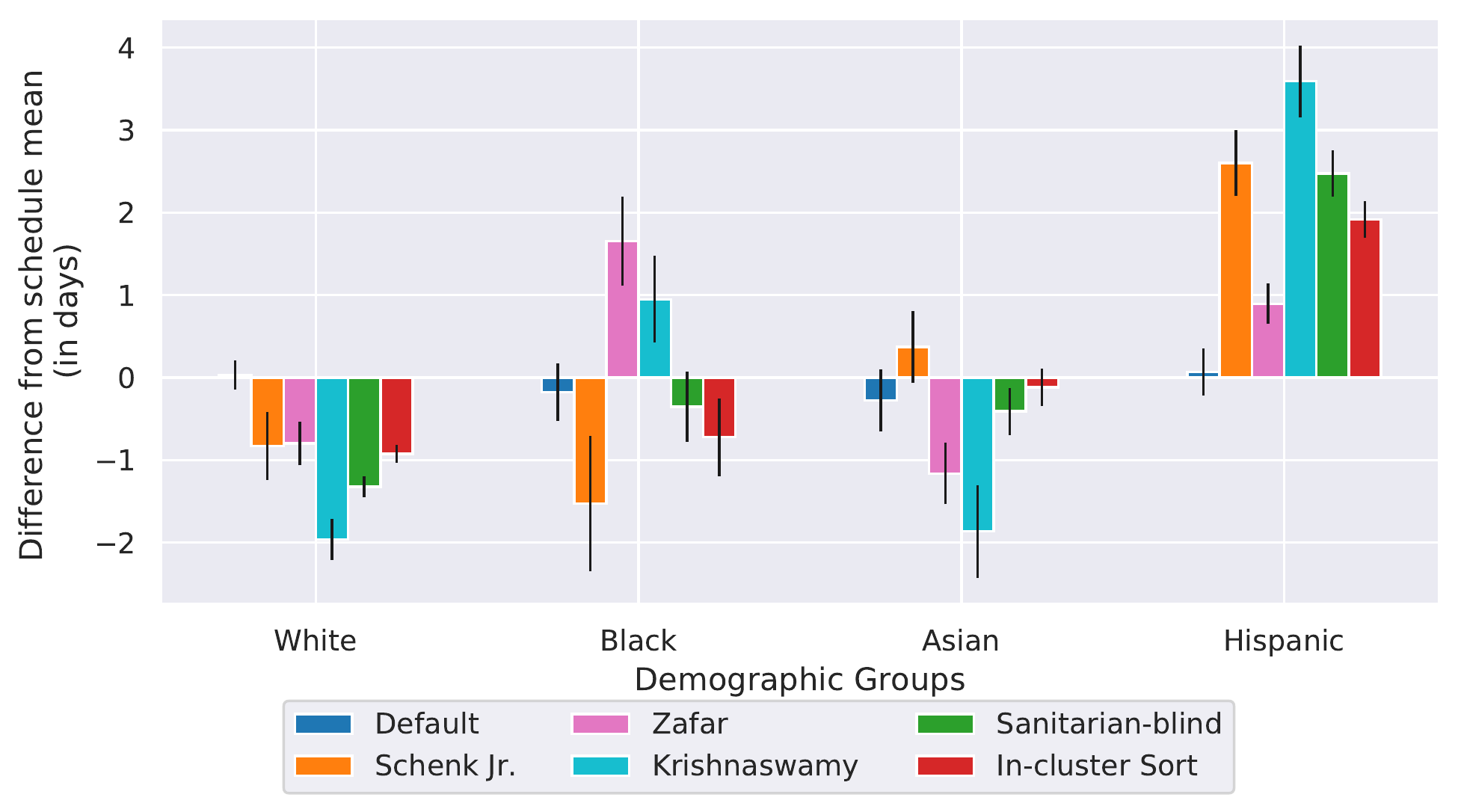}    
        \phantomsubcaption
        \label{subfig:postprocessing-race-dp}
    \end{minipage}
    \caption{The difference in detection (\cref{subfig:postprocessing-race}, left) and inspection (\cref{subfig:postprocessing-race-dp}, right) times for model-retraining approaches, explained in \Cref{sec:model-usage}.}
    \label{fig:app-postprocessing-race}
\end{figure}

\Cref{fig:app-inprocessing-race} shows the results for the approaches based on model retraining (explained in \cref{sec:model-retraining}) and \Cref{fig:app-postprocessing-race} for the model usage-based approaches (explained iin \cref{sec:model-retraining}).
The general trend supports our findings from our analysis for sanitarian cluster and city regions. 
The Schenk Jr.~schedule prioritizes the inspections in the regions that are heavily populated by the White and Black population and delays regions that are primarily Hispanic.
The improvements of model retraining and model usage-based approaches over the Schenk Jr.~schedule are less clear and overall the difference in detection and inspection times are substantially smaller than we saw with sanitarian or geographic groups, making it unclear how significant any unfairness is.

Overall, we view this analysis as essentially a null result. This is somewhat surprising given the high levels of racial segregation in Chicago; we had expected that geographic unfairness would translate to racial unfairness as well. One possibility is that the regions we use are too large or poorly chosen to capture the effects of segregation.  Another is that the approach we took to connect food inspections with demographics is inadequate.  We leave a fuller exploration of the challenge of quantifying such linkages to future work.

\subsection{Fairness Along Economic Lines}
\label{app:economic}

\begin{figure*}[h!]
    \centering
    \includegraphics[width=0.6\linewidth]{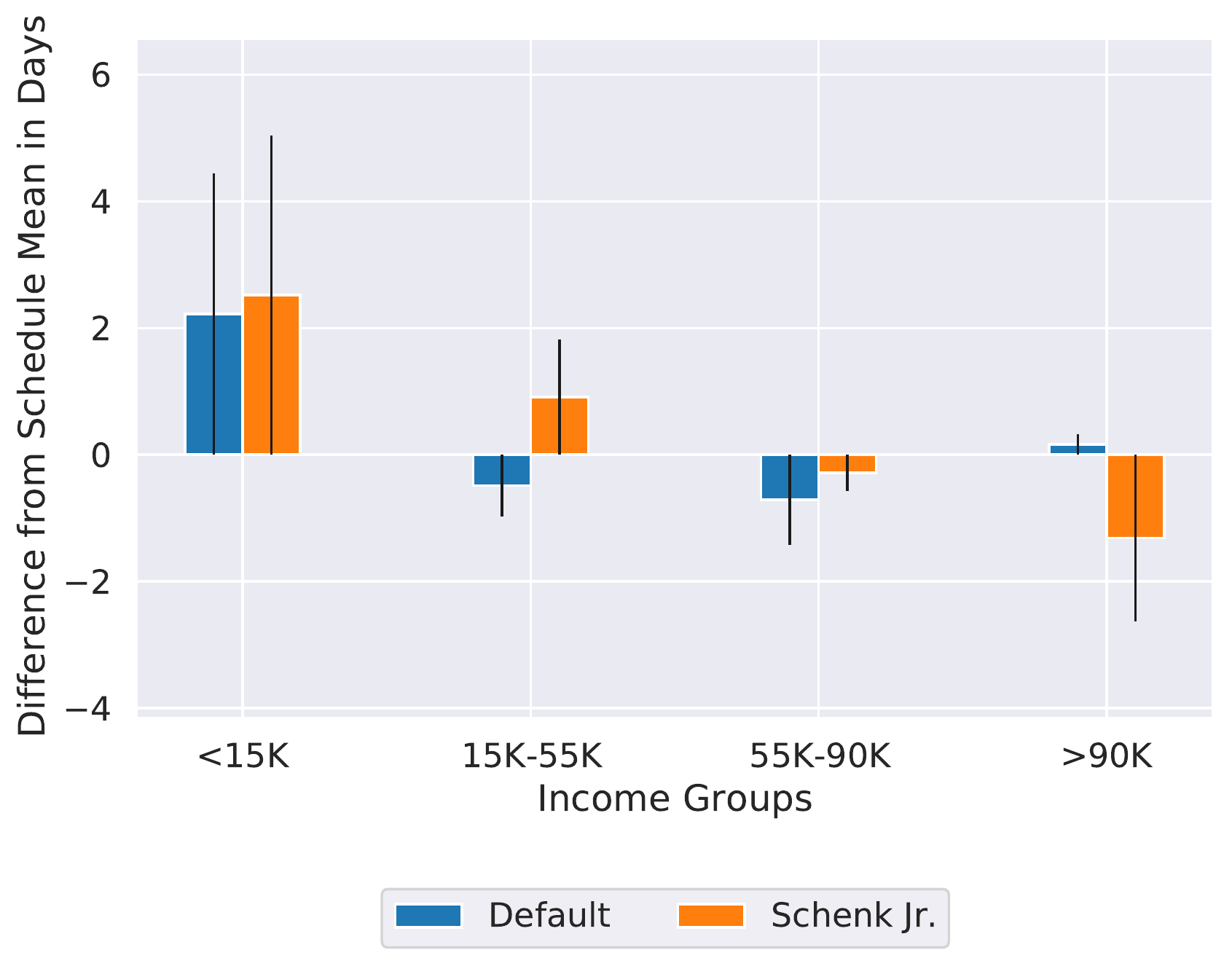}
    \caption{The times to detect violation under the schedules obtained by the original ordering and the Schenk Jr. model. The bars show the difference of the  detection times from the schedule mean by economic groups based on tax bracket with error bars giving the standard error. (Best viewed in color.)}
    \label{fig:income_federal_violation_time}
\end{figure*}

Another natural dimension to consider fairness across is economic groupings. We examined the fairness of the Default and Schenk schedules across economic groups using a similar approach to the demographic analysis in \ref{app:demographic}, leveraging the census data to understand how the orderings effected differ income groups. In \ref{fig:income_federal_violation_time}, the Default schedule is shown to be less beneficial to those with incomes of less than 15K and slightly beneficial or not beneficial to the other income groups. On the other hand the Schenk Jr.~schedule is also least beneficial to those in the lowest income group and gradually more beneficial to each following group. Specifically, it is the most beneficial to those with an income over 90K. 

Overall, as with our demographic analysis in \Cref{app:demographic} we view this analysis as essentially a null result. Though it is interesting to see a negative correlation between the Schenk Jr.~model's difference from schedule mean and the income groups, the magnitude of the standard error leads to inadequate conclusions. An interesting finding to explore in the future would be the small adverse effect of both the Default and the Schenk Jr. schedule on lower income groups.

Since, the population of customers may differ with the price range of the restaurant, we used the Yelp API to gather some additional data about the restaurants and we completed the same analysis as in \cref{sec:naive-analysis}. Our analysis is restricted as only about two thirds of inspections were conducted of establishments with Yelp profiles.

\begin{figure*}[h!]
    \centering
    \includegraphics[width=0.6\linewidth]{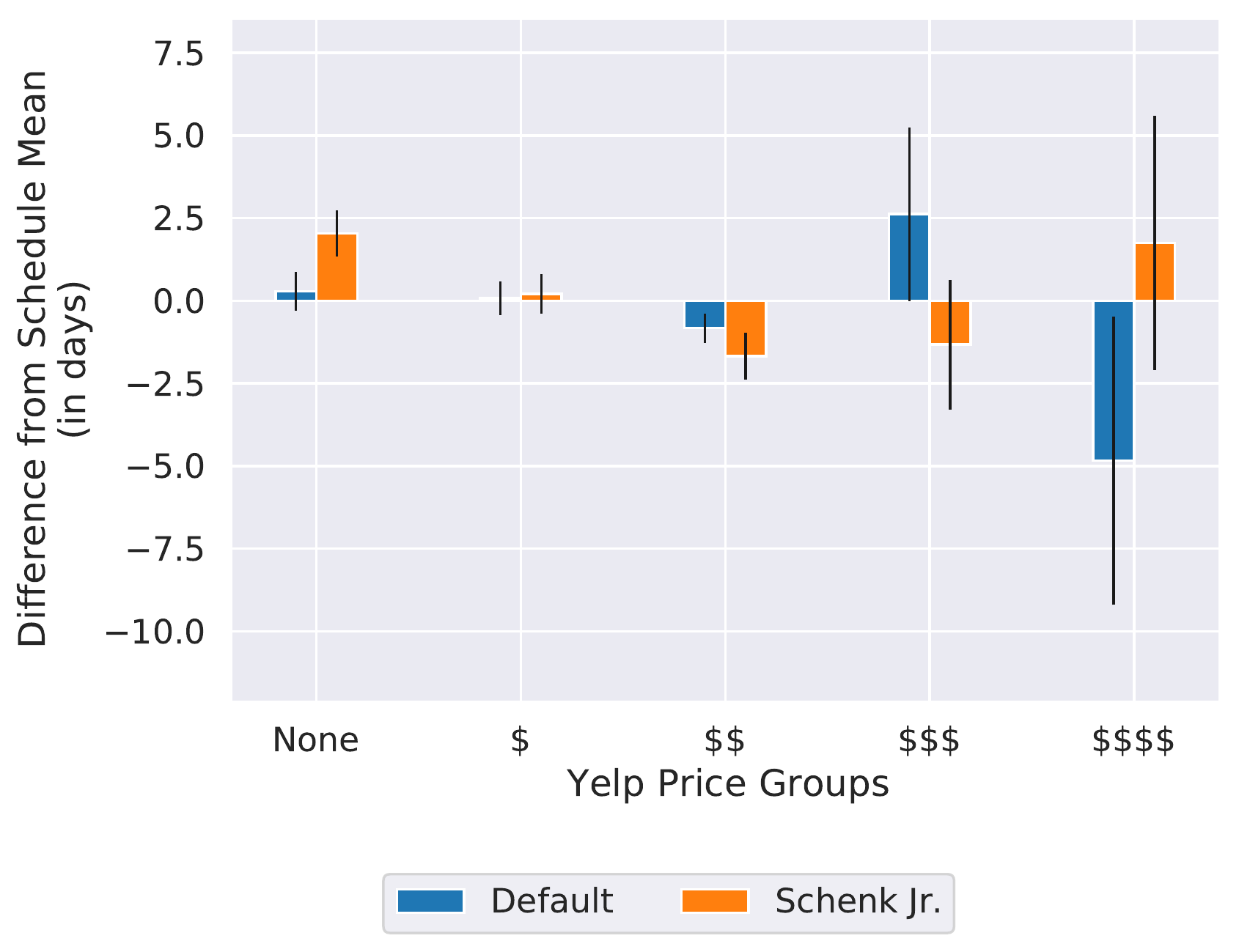}
    \caption{The times to detect violation under the schedules obtained by the original ordering and the Schenk Jr. model. The bars show the difference of the detection times from the schedule mean by the price level of the inspected restaurants with error bars giving the standard error. (Best viewed in color.)}
    \label{fig:yelp_violation_time}
\end{figure*}

The main attribute we looked at was the Yelp Price Level of the establishment being inspected.  \Cref{fig:yelp_violation_time} shows that the Schenk Jr. schedule seems to favor restaurants with more expensive offerings (\$\$\$\$). When interpreting \Cref{fig:yelp_violation_time}, it is important to keep in mind that not only are the effect sizes small, but the number of establishments in each price level group are not consistent, as shown in \Cref{tab:price-level}, leading to particularly large standard errors in the more expensive categories.

As with our demographic analysis,
exploring the connection between establishment based characteristics like price level, cuisine, and customer ratings with schedule order are all connections we leave for future work.


\begin{table}[ht]
    \centering
    \begin{tabular}{l c r}
    \hline
    \multicolumn{1}{l}{\textbf{Price Level}} & \multicolumn{1}{l}{\textbf{Dollars}} &
    \multicolumn{1}{l}{\textbf{Number of Inspections}} \\ \hline \hline
    No price data & NA & 5973                                            \\
    \$ &   $<10$ &   6896                                            \\
    \$\$ &   10-30   &   5282                                             \\
    \$\$\$ &   30-60   &   538                                               \\
    \$\$\$\$ &   $>60$ &   92                                               \\ \hline
    \end{tabular}
    \caption{Table showing the break down of inspections by the price level of the establishment.}
    \label{tab:price-level}
\end{table}

\end{document}